\pdfoutput=1
\documentclass[11pt]{article}

\usepackage[]{acl}

\usepackage{times}
\usepackage{latexsym}

\usepackage[T1]{fontenc}
\usepackage[utf8]{inputenc}

\usepackage{microtype}

\usepackage{graphicx}
\usepackage{multicol}
\usepackage{multirow}
\usepackage{booktabs}
\usepackage{amsmath}
\usepackage{upgreek}
\usepackage{footmisc}
\usepackage{tablefootnote}
\usepackage{footnote}
\usepackage{tabularx}
\usepackage{diagbox}

\title{RHO ($\rho$): Reducing Hallucination in Open-domain Dialogues with Knowledge Grounding}
\author{Ziwei Ji, Zihan Liu, Nayeon Lee, Tiezheng Yu, \\ 
\large{\textbf{Bryan Wilie, Min Zeng, Pascale Fung}} \\
       Center for Artificial Intelligence Research (CAiRE)\\
    Hong Kong University of Science and Technology\\
     \texttt{zjiad@connect.ust.hk, pascale@ece.ust.hk}}
\begin{document}
\maketitle
\begin{abstract}
% 1. hallucination in dialogue
% 2. The tolerance is low in kGD, the heterogeneity cause/challenging this issue.
% 3. our method improve the fusion and then reduce hallucination

% Benefiting from large-scale pre-trained language models, knowledge-grounded dialogue (KGD) systems achieve impressive performance and are able to generate fluent responses.
% However, the neural models are prone to produce hallucinated responses, which hinder the application scenarios of these systems. 
% Dialogue systems have made promising achievements, yet they are still prone to generating hallucinated responses that contradict or cannot be verified with the input.
Dialogue systems can leverage large pre-trained language models and knowledge to generate fluent and informative responses. However, these models are still prone to produce hallucinated responses not supported by the input source, which greatly hinders their application.
% zihan: Dialogue systems can leverage context-relevant knowledge to generate informative responses. However, these models are still prone to produce factually incorrect responses, which could greatly hurt the model quality and even misguide users to use the wrong information.
% hallucination definition, assume they know nothing 
% This issue hinders the successful and safe deployment of knowledge-grounded dialogue systems. 
The heterogeneity between external knowledge and dialogue context challenges representation learning and source integration, which further contributes to unfaithfulness.
% /potentially
% poses a challenge to their fusion/interaction in knowledge-grounded dialogue and further cause this divergence.
% representation learning, comprehension, and generation in KDG task.
% To generate more faithful responses, this paper presents \textbf{RHO}, a knowledge-grounded encoder-decoder model with the dialogue contexts grounded by the representations of linked entities and relation predicates from a knowledge sub-graph related to the contexts.
To handle this challenge and generate more faithful responses, this paper presents \textbf{RHO} ($\rho$) utilizing the representations of linked entities and relation predicates from a knowledge graph (KG).
% % The representations of dialogue contexts are grounded by the representations of linked entities and relation predicates from a context-related knowledge sub-graph.
% a knowledge-grounded encoder-decoder
% context-related knowledge sub-graph??
% We implement the knowledge grounding process by (1) local grounding, where an entity or relation predicate only receives/utilizes its own knowledge graph (KG) embedding; and (2) global grounding, where each entity or relation predicate attentively learns the knowledge representation from the entire sub-graph stored in a memory bank to equip \textbf{RHO} with multi-hop reasoning abilities.
We propose (1) {local knowledge grounding} to combine textual embeddings with the corresponding KG embeddings; and (2) {global knowledge grounding} to equip \textbf{RHO} with multi-hop reasoning abilities via the attention mechanism.
In addition, we devise a response re-ranking technique based on walks over KG sub-graphs for better conversational reasoning.
%with explainable conversation reasoning.
Experimental results on OpenDialKG~\cite{moon2019opendialkg} show that our approach significantly outperforms state-of-the-art methods on both automatic and human evaluation by a large margin, especially in hallucination reduction (17.54\% in FeQA~\cite{durmus-etal-2020-feqa}).\footnote{The source code can be obtained from~\url{https://github.com/ziweiji/RHO} }
\end{abstract}

\section{Introduction}
\label{sec:introduction}
\begin{figure}[!t]
 \centering
 \includegraphics[width=1\linewidth]{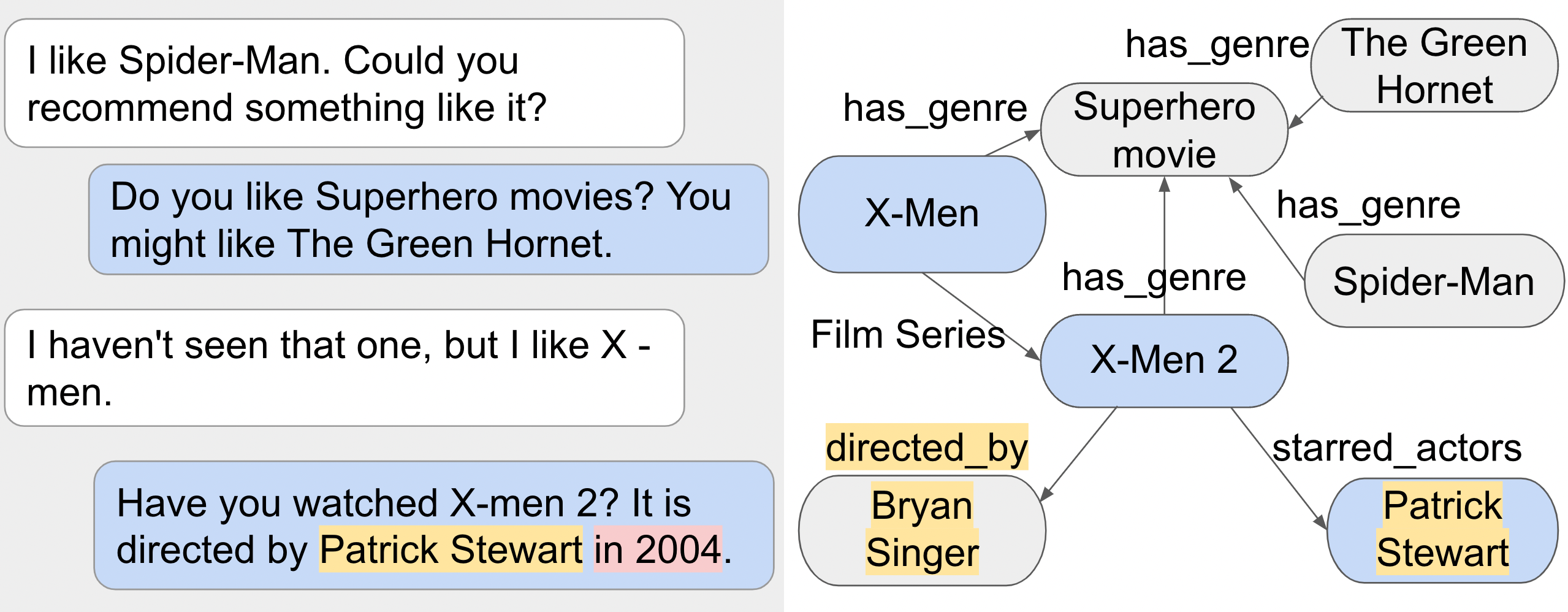}
  \caption{An example of hallucinated KGD. 
  Based on the dialogue history and the KG, the system is expected to generate a response.
   In the response, \texttt{``Patrick Stewart''} contradicts the fact \texttt{<X-Men\ 2, directed\_by, Bryan\ Singer>}, which is a case of intrinsic hallucination (in yellow); \texttt{``in 2004''} cannot be verified based on the given information, which is a case of extrinsic hallucination (in pink).}
  \label{fig:example}
    \vspace{-1em}
\end{figure}

% Para 1: Dialogue systems -> important applications in NLP -> generate appreciate responses for users -> key problem (faithfulness)
% An open-domain dialogue system aims to automatically interacts with humans with generated responses and provides users with an engaging experience.
 %generates responses according to the users' utterances 
%  to generate meaningful and informative responses
% the advancement of KGD models
An open-domain dialogue system aims to automatically interact with humans with sensible and informative responses.
To produce such responses, knowledge-grounded dialogue (KGD) systems are established, which leverage external knowledge such as knowledge graphs (KGs)~\cite{yu2022survey,zhao2020knowledge}.
Despite impressive progress in general linguistic performance, KGD systems are still prone to the hallucination problem in which the generated response is nonsensical or unfaithful to dialogue history or external knowledge~\cite{ji2022survey,roller2021recipes,mielke2022reducing}. 
% the provided contexts, i.e.,
Two types of hallucinations may exist: intrinsic hallucination (the generated response is contradicted by the dialogue history or the external knowledge) and extrinsic hallucination (the generated response is hard to be verified with the dialogue history and external knowledge)~\citep{ji2022survey,dziri-etal-2021-neural}. 
As the example in Figure~\ref{fig:example}, this issue undermines dialogue systems' performance or raises safety concerns in real-world applications.
For instance, the recently emerged foundation model ChatGPT suffers from this hallucination problem~\citep{openai_2023}, especially extrinsic hallucinations. OpenAI currently uses huge amounts of human feedback to fix many ChatGPT errors, which is labor-intensive. It would be beneficial to reduce such errors automatically in advance.%the first place

% ~\footnote{\url{https://openai.com/blog/chatgpt/}}
% ChatGPT sometimes writes plausible-sounding but incorrect or nonsensical answers. Please refer to Appendix~\ref{appendix sec: chatgpt example} for examples.

% This issue undermines the performance of conversational AI in real-world applications and has raised much research interest~\cite{shuster2021retrieval}.  

%  maximizing knowledgeability while retaining conversational ability.  
%  However, the input text alone often provides limited knowledge

% Para 2-1. How to improve faithfulness -> kg-grounded approach for response generation -> brief summarization of existing works to reduce hallucination

% %the cause of hallucination in KGD (challenges of KGD itself)
The heterogeneity between external knowledge and textual dialogue content makes it challenging for neural response generation models to learn the representation and correlation in the input source~\citep{li-etal-2022-knowledge,zhang2019ernie,liu2020k,ji2022survey}. This challenge could further result in a hallucinated generation that deviates from the source. % diverge
% /attention
%The gap / format  between the textual dialogue history and external structured KG bring challenges to representation learning, comprehension and generation in KDG task.
% work about hallucination -> the unsolve problem
Previous studies have shown that the hallucination problem in KGD can be mitigated by retrieved knowledge augmentation~\citep{shuster2021retrieval}, control code~\citep{rashkin2021increasing, wu2021controllable}, and response post-processing~\citep{dziri-etal-2021-neural}.
% For example,~\citet{shuster2021retrieval} conduct a comprehensive study on how augmenting dialogue generation with the retrieved knowledge helps discourage hallucinations.
% \citet{rashkin2021increasing, wu2021controllable} control the extent to which the model focuses on external knowledge during response generation. 
% Neural Path Hunter (NPH)~\citep{dziri-etal-2021-neural} detects hallucinated entities with a hallucination critic in the generated responses and amends them with entities retrieved from the KG. 
% "We observed that the attention-based generation are often adopted in the DGDS." from A Survey of Document Grounded Dialogue Systems (DGDS)
%Existing works tend to employ encoder-decoder or fully decoder architectures for the task~\cite{yu2022survey,xu2020knowledge, liu-etal-2018-knowledge, zhou2018commonsense,meng2021initiative}.
% Para 2-2. Key drawbacks and challenges. unsolved issues
However, these works do not emphasize handling the discrepancy between lexical dialogue and structured knowledge information for the harmony of their fusion. The interaction mechanism between external knowledge and dialogue context should also be clarified.
% % We don't focus on the explainability
% % The ``black-box'' nature of the neural response generation models (such as encoder-decoder or fully decoder models) make it hard for the models to understand the semantics of key entities and relations~\cite{yu2022survey,xu2020knowledge, liu-etal-2018-knowledge, zhou2018commonsense,meng2021initiative}).
% % It is also unclear how the dialogue history and the question should interact with external knowledge in the process to improve the faithfulness of responses.

%Thus, we leverage the graph representation techniques and propose a novel knowledge injection mechanism.

\begin{figure*}[!ht]
 \centering
 \includegraphics[width=1\linewidth]{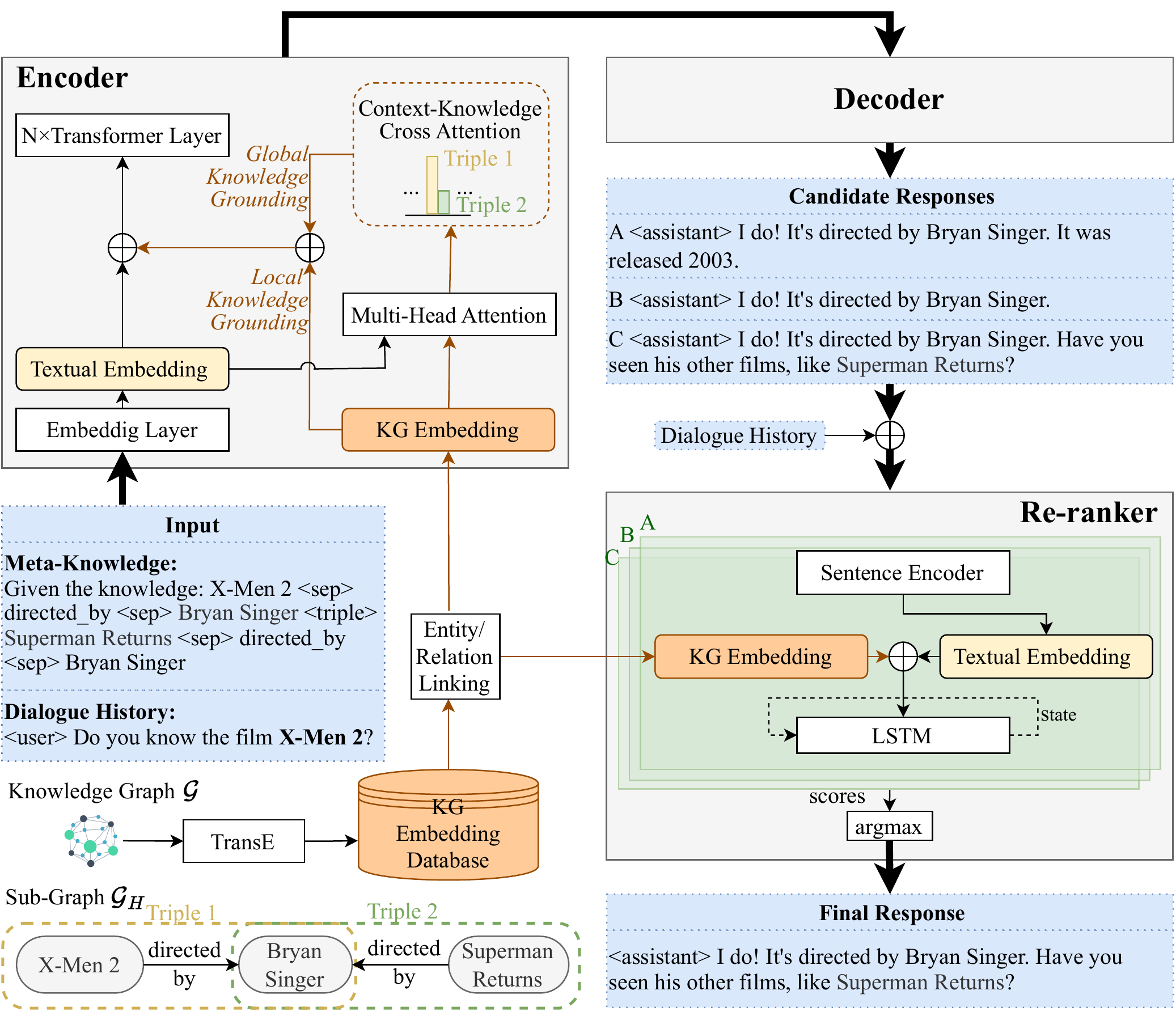}

  \caption{
  The overview of the proposed \textbf{RHO} framework.
%   In dataset preprocess:
    The input follows a knowledgeable task guidance template, including meta-knowledge and dialogue history (\S \ref{sec:guidance}).
    % "Given the knowledge: Great Expectations <sep> written by <sep> Charles Dickens <triple> A Christmas Carol <sep> written by <sep> Charles Dickens <user> Do you know the book Great Expectations?"
%   In encoder:
  To facilitate knowledge grounding, we first employ TransE to gain KG embeddings.
  %   local:
  For local knowledge grounding, we adapt entity/relation linking to recognize mentions in the dialogue context. The corresponding KG embeddings are locally fused into textual embeddings (\S \ref{sec:local}).
  % In this example, "Great Expectations" in dialogue is linked to the corresponding entity in KG.
    %   global:
  For global knowledge grounding, we aggregate the entire knowledge sub-graph in a memory bank via the attention mechanism so that the textual embeddings receive all context-related knowledge with emphasis (\S \ref{sec:global}). 
  %For token "Great", Triple1 gain higher context-knowledge cross attention score than Triple2.
  After that, the encoder-decoder model generates several candidate responses.
%   A "<assistant> I do! It's written by Charles Dickens. It was released 1838."; B "<assistant> I do! It's written by Charles Dickens."; and C "<assistant> I do! It's written by Charles Dickens. Have you read any of his books, like A Christmas Carol?".
   %   re-rank:
  During post-processing, a re-ranker trained by traversal over a knowledge sub-graph conditioned on the dialogue context selects the most faithful response as the final output (\S \ref{sec:re-rank}).
  %C because of the highest probability of predicting the given/golden knowledge path 
%   Finally, the re-ranker selects response C because it gains the highest probability of predicting our golden knowledge path.
  }
  \label{fig:system}
  \vspace{-1em}
\end{figure*}

In order to address this issue and take full advantage of lexical and knowledge information, we present \textbf{RHO} ($\rho$)~\footnote{\textbf{RHO} is short for \textbf{R}educing \textbf{H}allucination in \textbf{O}pen-domain dialogue systems.} for faithful open-domain dialogue response generation with enhanced knowledge grounding and re-ranking.
% from external knowledge to dialogue contexts. 
A high-level framework is illustrated in Figure~\ref{fig:system}.
Specifically, \textbf{RHO} first learns the structured embeddings of all entities and relation predicates from the KG and links their mentions in the dialogue context to the KG.
%a sub-graph into the dialogue context based on entity and relation predicate linking.
% In the knowledge-grounded encoder-decoder model
In the encoder-decoder model, the representations of all the linked entities and relations are grounded by KG embeddings, both~\emph{locally} and~\emph{globally}.
Here, \emph{local knowledge grounding} refers to the process where an entity or relation predicate receives and fuses its KG embedding only. 
While in~\emph{global knowledge grounding}, each entity or relation predicate attentively learns the knowledge from the entire sub-graph stored in a memory bank~\citep{vaswani2017attention}, which assigns dynamic weights of each triple equipping \textbf{RHO} with multi-hop reasoning abilities.
These two knowledge groundings help the model effectively encode and inject the knowledge information from context-related sub-graph with proper attention.
% As injecting too much knowledge may cause the knowledge noise problem~\cite{liu2020k}, we further design a (TBD) technique to selectively enforce knowledge grounding across different layers of our \textbf{RHO} model.
In addition, we re-rank the generated responses according to the hallucination degree.
% , utilizing conversational reasoning and walks over KGs.
This technique utilizes conversational reasoning to enforce the whole conversation to follow the knowledge traverses throughout KG.

% rich local and global knowledge injection, prompt for task guidance.
% for knowledge, \textbf{RHO}
% for prompts, \textbf{RHO}
% (Need some fancy name here)

% Para 4. Brief summary of results. Outperform \textbf{RHO} by a large margin (xx -> xx in terms of \textbf{RHO}). Detailed analysis: better coverage of entities and relations in KG -> more faithful
In the experiments, we show that \textbf{RHO} outperforms state-of-the-art (SOTA)~\citep{dziri-etal-2021-neural} on the OpenDialKG~\cite{moon2019opendialkg} dataset by a large margin: improving 17.54\% in FeQA~\cite{durmus-etal-2020-feqa}, and reducing 32.93\% hallucinations according to human evaluation.
% \% in BLEU4~\cite{papineni2002bleu} and
% We also achieve better results compared to the stronger baseline which replaces GPT2 with BART as its backbone.
In particular, the responses have a broader coverage of entities and relations in the KG, demonstrating higher faithfulness of responses.
The quantitative and qualitative analysis further shows its effectiveness in reducing hallucination while not sacrificing conversational abilities.
% Para 5. Summary of contributions
In summary, the major contributions of this work are threefold:
\begin{itemize}
\vspace{-0.5em}
\setlength{\itemsep}{0pt}
\setlength{\parsep}{0pt}
\setlength{\parskip}{0pt}
%偏task 不写结果 key insight
\item We propose the \textbf{RHO} model, which leverages the structured knowledge in KGs to mitigate the hallucination problem in dialogue response generation. 
%偏technique
\item To improve faithfulness, we introduce local and global knowledge grounding techniques (from a context-related knowledge sub-graph) into dialogue generation and further utilize a conversational reasoning model to re-rank the generated responses. 
% Notably, the global knowledge grounding adopts the attention mechanism to draw global dependencies between the dialogue history and memory bank.

% context-related knowledge sub-graph stored in a
\item We conduct a thorough faithfulness analysis via automatic and human evaluation, and empirically demonstrate that \textbf{RHO} substantially reduces intrinsic and extrinsic hallucinations in the KGD generation task.
\end{itemize}

\section{Related Work}
\label{sec: related}
% Part I. kg-grounded response generation for dialogues. Specifically, methods for hallucination reduction (2/3)
\subsection{Hallucination Reduction in KGD}
%from industry and academia 
Researchers have been devoted to reducing hallucination in open-domain dialogue systems incorporating external knowledge.
% ~\cite{ma2020survey}
Neural Path Hunter (NPH)~\citep{dziri-etal-2021-neural} leverages a hallucination critic and retrieves faithful entities by a query signal propagated over a sub-graph in the refinement stage. 
%In this way, the amended responses are semantically supported by the provided KG.
\citet{shuster2021retrieval} explore various neural-retrieval-in-the-loop architectures where a retriever is introduced for knowledge selection.
% In particular, they employ Poly-encoder Transformers~\cite{humeau2019poly} to refine the scoring mechanism.
%of candidate documents,
%employ end-to-end-trained retrievers in the 
%Fusion-in-Decoder~\cite{izacard2021leveraging}
%adopts an iterative retrieval strategy and employs a retrieval scheme based on dialogue turns.
\citet{rashkin2021increasing} propose an faithfulness control code in decoding using re-sampling techniques.
\citet{wu2021controllable} define a control mechanism with lexical control phrases and inductive attention where potentially uninformative attention links are removed.
% by injecting pre-established structural information between the control phrase and relevant grounding.
Our work improves the fusion and interaction between external knowledge and dialogue context via various knowledge groundings and reasoning techniques, further reducing hallucination.

\subsection{KG Enhanced Dialogue Generation}
 KGs convey large amounts of structured knowledge, which can help to improve dialogue systems' performance in informativeness~\cite{tuan-etal-2019-dykgchat} and empathy~\citep{li2020knowledge}.
For open-domain dialogue generation, \citet{liu2019knowledge} unify knowledge triples and texts as a graph, and conduct multi-hop reasoning for explainability.
% introduce two RNNs for computing the representation of the knowledge and conversation separately, and then combine three probability distributions to produce the next token. % an augmented KG and a reasoning-based knowledge selector.
\citet{xu2020knowledgegraph} propose a proactive dialogue generation method based on agnostic meta-learning considering the limited number 
of KGs.
\citet{kumar-etal-2020-amused} learn unified representations by training syntactic graph convolution networks, knowledge, and memory module with triplet loss.
% For open-domain commonsense-aware dialogue generation,
\citet{xu2022open,zhou2018commonsense,zhang-etal-2020-grounded} explore and demonstrate how commonsense KG facilitates language generation in dialogue systems.
%  DMKCM~\cite{xu2022open} dynamically fuse virtual knowledge base (KB) and commonsense KG to generate better responses.
% \cite{zhou2018commonsense} through a dynamic graph attention mechanism
% ConceptFlow~\cite{zhang-etal-2020-grounded} leverages commonsense KGs to model conversation flows explicitly in the latent concept space for generating informative and meaningful responses.
Besides, \citet{yang-etal-2020-graphdialog, rony2022dialokg, chaudhuri2021grounding} are committed to incorporating KG into task-oriented dialogue models. 
 % To incorporate KG into a task-oriented dialogue model, \citet{yang-etal-2020-graphdialog} leverage a recurrent cell architecture and a multi-hop reasoning mechanism 
 % \citet{rony2022dialokg} leverages knowledge embedding technique and KG-weighted attention masking in task-oriented dialogue model.
 % \citet{chaudhuri2021grounding} propose a decoding method that integrates KG into a pre-trained transformer model via Graph Laplacian in task-oriented dialogue generation.
 % %%%%%%%%%%%%%%%%%%%%%%%
% Other scenarios such as empathetic conversation~\citep{li2020knowledge}
% KEMP~\citep{li2020knowledge} constructs an emotional context graph and learns emotional context representations to generate empathetic responses.
% DyKgChat~\cite{tuan-etal-2019-dykgchat} proposes a task of applying dynamic KGs in conversation models to demonstrate their zero-shot adaptability.
% These works do not explore the hallucination problem
Different from the above literature, our work employs the factoid knowledge paths from KG to improve the \textit{faithfulness} of open-domain dialogue systems.

\begin{figure*}[!ht]
 \centering
 \includegraphics[width=1\linewidth]{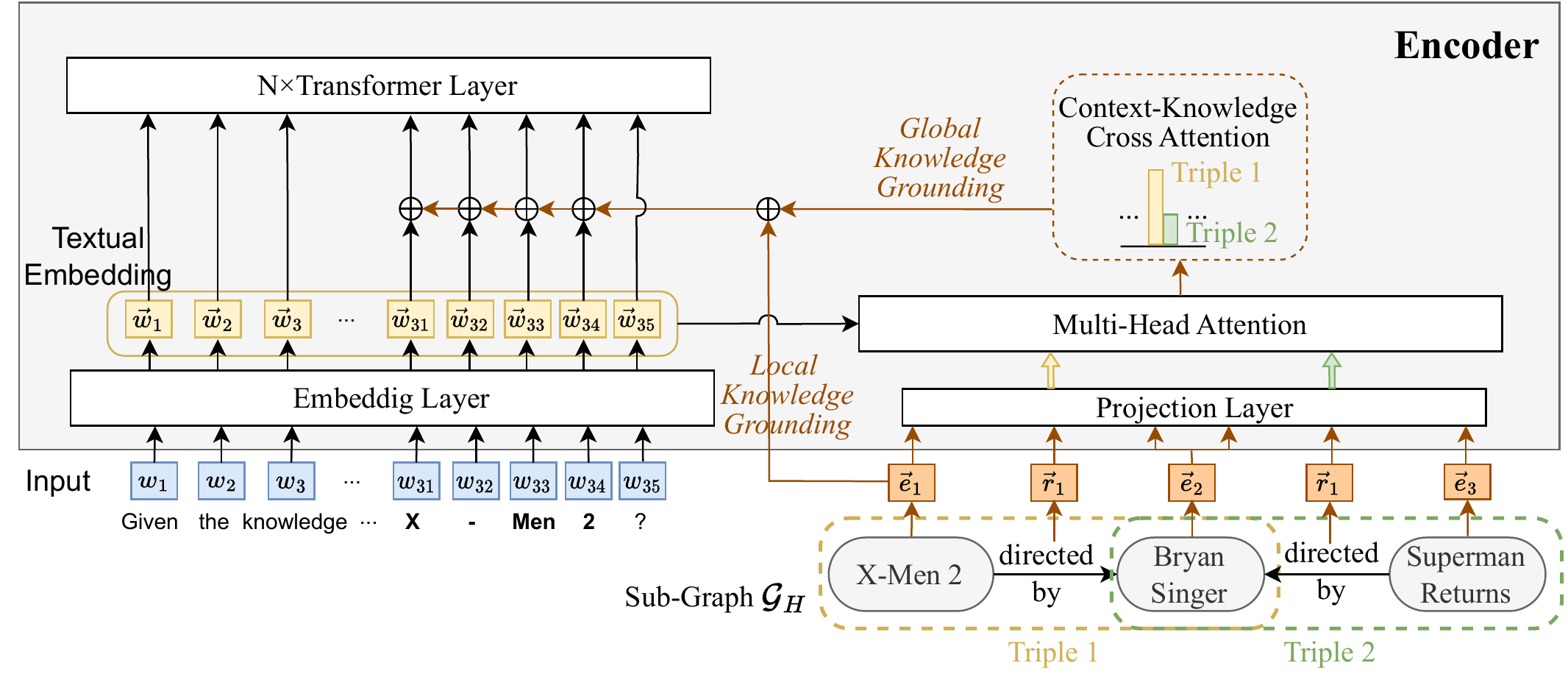}
  \caption{The diagram of the encoder with proposed local and global knowledge groundings. }
  \label{fig:encoder}
  \vspace{-1em}
\end{figure*}

\section{Methodology}
% Part II. overview of proposed approach
% why do it in this way? not only how. need to add the motivation
In this section, we begin with a brief introduction to our KGD task. Then, the detailed techniques of \textbf{RHO} are presented. Please refer to Figure~\ref{fig:system} for an overview of our approach.
\textbf{RHO} incorporates both textual and structured information from external KG into 
% language representations
dialogue system via knowledgeable task guidance (\S~\ref{sec:guidance}) and enhanced knowledge groundings (\S~\ref{sec:local}, \S~\ref{sec:global}).
Specifically, we introduce local token-level knowledge grounding in \S~\ref{sec:local} and global grounding to provide a comprehensive view and the multi-hop reasoning ability in \S~\ref{sec:global}.
% In addition to grounding a token locally on the corresponding knowledge, we also introduce global grounding to provide a comprehensive view and the multi-hop reasoning ability.
% % These techniques help our model to effectively encode the heterogeneous sources
To effectively encode the heterogeneous sources, we sum textual embedding, local, and global grounded embedding. Figure~\ref{fig:encoder} is a more detailed diagram of the encoder with proposed knowledge groundings.
In addition, during the post-processing stage, we incorporate a re-ranking technique that rewards generation candidates with low hallucination levels; this technique is proposed based on the hypothesis that the faithful response can be reasoned backward to the source (\S~\ref{sec:re-rank}).
% a re-ranking technique according to the hallucination level is introduced in the post-processing stage with the hypothesis that the faithful response can be reasoned backward to the source (\S~\ref{sec:re-rank}).
% intuition/
% , which also achieves more explainable conversation reasoning.
We implement our model based on BART~\citep{lewis2020bart} architecture, and please refer to Appendix~\ref{appendix sec: implement} for details.

\subsection{Task Definition}
In the response generation task in dialogue systems, each data sample consists of a dialogue history $H$, including a set of utterances $U$ conducted by humans and agents interactively.
The goal of the response generation model in the dialogue system is to learn to generate a proper response $R$ based on the dialogue history $H$.

The response generation for KGD task is a special case of the above task that takes a multi-relational KG as an additional input; Multi-relational KG is a directed graph $\mathcal{G}$ formulated by a collection of triples, denoted as $T=\left \langle \texttt{[SBJ],[PRE],[OBJ]} \right \rangle$, as additional input.
Here, \texttt{[SBJ],[OBJ]} are subject and object entities, and \texttt{[PRE]} is the relation predicate $r$ indicating the relationship between the subject and object entities $e$.
The goal of KGD is to generate a faithful response $R$ based on the history $H$ and a knowledge sub-graph $\mathcal{G}_H$,  which is the subset of the entire KG $\mathcal{G}$ with triples semantically related to the dialogue history $H$.
Our task is also in line with the previous works~\citep{zhou2021earl,dziri-etal-2021-neural}.
Figure~\ref{fig:system} has an illustrated example.

\subsection{Input with Knowledgeable Task Guidance}
\label{sec:guidance}
%two types of content-matching constraints~\cite{wang2020towards}:
%1. an explicit entity-level matching scheme
%2. a latent-representation level matching constraint}
A naive approach of input construction from raw data samples for language models is simply concatenating all triples in $\mathcal{G_H}$ with the dialogue history $H$\footnote{Similar to the baseline approach in \citet{dziri-etal-2021-neural}.}.
However, there is a lack of guidance to specifically excavate the model's innate ability to handle the KGD task \citep{brown2020language}.
%downstream task in PLM. %  
Inspired by~\citet{raffel2020exploring}, we design a prompt to guide the PLM for KGD and convert the structured $\mathcal{G_H}$ into textual information.
% we augment inputs to guide the PLM for KGD and fully encode the structured knowledge in $\mathcal{G_H}$.
Here, we linearize the triples in $\mathcal{G_H}$ into texts (treated as meta-knowledge) and cooperate the dialogue history utterances $U$ with the following template: ``\texttt{Given the knowledge: [SBJ]$_1$ <sep> [PRE]$_1$ <sep> [OBJ]$_1$ <triple> [SBJ]$_2$ <sep> [PRE]$_2$ <sep> [OBJ]$_2$ <triple> $\cdots$ <user> U$_1$ <assistant> U$_2$ $\cdots$}'', where \texttt{<sep>}, \texttt{<triple>}, \texttt{<user>}, and \texttt{<assistant>} are special markers.
% Figure~\ref{fig:system} has an illustrated example.

%The experiments also prove that the designed task-specific template provides proper contextual task knowledge to large PLMs in order to generate better responses.

\subsection{Local Knowledge Grounding}
\label{sec:local}
% Entire KG -> KG embeddings
% Entity and relation linking
% Knowledge injection by embedding fusion -> enrich the semantics of entities and relation predicates themselves

%Given a input token sequence $\left \{ w_1,...,w_n \right \}$, the original token embedding gained from textual PLM encoder can be expressed as $\mathbf{H_{text}} = Encoder(\left \{ w_1,...,w_n \right \})$.
% why incorporate KG embedding to language model
% self-attention between prefix and dialogue history??? 
% shall I add the following reasons why incorporate KG embedding to PLM to introduction?}

Although $\mathcal{G_H}$ is converted and injected as additional input (\S~\ref{sec:guidance}), the model using textual information only cannot effectively handle the semantics of KGs which are typically sparse and complex in form~\cite{petroni-etal-2019-language,logan-etal-2019-baracks}.
% Although the triples in $\mathcal{G_H}$ are translated into plain texts as a part of the input, the model using textual information only cannot well handle the semantics of the specific entities and their relations, which are typically sparse and in complex forms~\cite{petroni-etal-2019-language,logan-etal-2019-baracks}.
Therefore, we ground the language representations with KG to take full advantage of lexical and structured knowledge information simultaneously.
% In \textbf{RHO}, we enforce local knowledge grounding by means of heterogeneous knowledge fusion.

% without explicit knowledge grounding on specific entities and relation predicates,

%To avoid knowledge noise problem~\cite{liu2020k}, we only integrate the graph embedding into the tokens belonging to the KG. 

% The linking/alignment result are $\mathbf{\mathcal{E}}=\left \{ e_1,...,e_I \right \}$ and $\mathbf{\mathcal{R}}=\left \{ r_1,...,r_J \right \}$
%One entity $e\in\mathcal{E}$ or relation $r \in\mathcal{R}$ includes a token sequence $\{w_{i},\cdots,w_{j}\}$.
% Consider one entity $e\in\mathcal{E}$ the token sequence of the entity mention in \textbf{RHO} is $\{w^{(e)}_i,\cdots,w^{(e)}_j\}$.

During pre-processing, we obtain the collections of linked mentions of entities ($\mathcal{E}_H$) and relations ($\mathcal{R}_H$) from the dialogue history $H$ and their KG embeddings as follows:
\begin{enumerate}
\vspace{-0.5em}
\setlength{\itemsep}{0pt}
\setlength{\parsep}{0pt}
\setlength{\parskip}{0pt}
    \item Identify entity mention $e_m$ that appears in dialogue history $H$ that can be linked to an entity $e$ in the sub-graph $\mathcal{G_H}$. We utilise an open-source linking tool named FLAIR~\cite{akbik2019flair}. 
    
    \item Since relations connecting entities are crucial in knowledge reasoning for PLMs~\cite{labutov-etal-2018-multi,feng-etal-2020-scalable}, we also link the relation mention $r_m$ in $H$ to the relation predicate $r$ in $\mathcal{G}_H$. 
    % \item Hence, for the entire dialogue history $H$, we have the collections of linked mentions of entities and relation predicates, represented as $\mathcal{E}_H$ and $\mathcal{R}_H$.

    \item We employ TransE~\citep{bordes2013translating} to learn the KG embeddings of entities ($\vec{e}_\mathcal{G}$) and relation predicates ($\vec{r}_\mathcal{G}$) from the entire $\mathcal{G}$~\footnotemark{}.
\vspace{-0.5em}
\end{enumerate}

Then, we obtain a \emph{locally grounded} token embedding $\vec{w}_{local}$ for an arbitrary non-special token $w$ in $H$ as follows:
% For an arbitrary non-special token $w$ in $H$,
% the~\emph{locally grounded} token embedding $\vec{w}_{local}$ is then defined as follows:
% %To fuse heterogeneous information, we inject the knowledge graph embedding from TransE~\citep{bordes2013translating} into the underlying layers of the semantic model. 
% %Since one knowledge triple is restricted to particular tokens, we call it local knowledge injection. 
% %For any entity $e_i$ in $\mathcal{E}$ and any relation $r_j$, it is expressed as follows:
\begin{equation}
\setlength\abovedisplayskip{3pt}%shrink space
\setlength\belowdisplayskip{3pt}
\vec{w}_{local}=\left\{\begin{matrix}
M(\vec{e}_\mathcal{G})  & \texttt{substr}(w, \mathcal{E}_H)\\ 
M(\vec{r}_\mathcal{G})  & \texttt{substr}(w, \mathcal{R}_H)\\ 
\vec{0} & \text{otherwise} \end{matrix}\right.
\label{equ: local}
\end{equation}
where $M(\cdot)$ transforms the space from the KG embeddings to the PLM token embeddings.
A typical way of implementing $M(\cdot)$ is through a mapping matrix. Specifically, if $dim(\vec{w})=dim(\vec{e}_\mathcal{G})$, $M(\cdot)$ can be further simplified as an identity mapping. 
$\texttt{substr}(w, \mathcal{E}_H)$ is a Boolean indicator that returns true if the current token $w$ is a sub-string of any $e_m$ in $\mathcal{E}_H$.
% specific entity or relation names
%For example, in Figure~\ref{fig:system}, the representations of all tokens in \textit{``Great Expectations''} are grounded by its KG embedding.
This way, tokens related to specific entities or relation predicates can be grounded by their respective KG embeddings by fusing $\vec{w}_{local}$ into the vanilla token embedding $\vec{w}$.
We regard this approach to be~\emph{local} as 
$\vec{w}_{local}$ is only related to the KG embedding of the corresponding node.
\footnotetext{We employ TransE via OpenKE~\citep{han2018openke} as our underlying KG representation learning algorithm due to its effectiveness and simplicity. We have also experimented with some recent algorithms and observed no improvement in the performance (refer to Appendix~\ref{appendix sec: kg repres} for details).}
%in $\mathcal{G}$. %in certain regions,
%without considering the overall view of the sub-graph $\mathcal{G}_H$.
%without the holistic/comprehensive/overall view/representation/information? of the entire sub-graph $\mathcal{G}_H$.

As in Figure~\ref{fig:system} and Figure~\ref{fig:encoder}, the tokens \texttt{"X", "-", "Men", "2"} in dialogue history are linked to the entity \texttt{"X-Men 2"} in KG. Then, we take the corresponding KG embedding from the database gained by TransE as the local knowledge grounding which will fuse into these tokens' textual embeddings.

%For token $w_n$, the local knowledge embedding $\mathbf{H_{local,n}}$ is the sum of $\mathbf{H_{local,n,i,j}}$ over all entities in $\mathcal{E}$ and all relations in $\mathcal{R}$.
%\begin{equation}
% \setlength\abovedisplayskip{3pt}%shrink space
% \setlength\belowdisplayskip{3pt}
%H_{local,n}=\sum _{i}^{I}\sum _{j}^{J} %H_{local,n,i,j}
% \vspace{-0.5em}
%\end{equation}

\subsection{Global Knowledge Grounding}
\label{sec:global}
% Subgragh-based Knowledge Injection
% KG subgraph retrieval??
% Construction of memory banks??
% Knowledge injection by attentive knowledge retrieval -> enrich the semantics considering the entire subgraph
% Focusing only on the fact that current token align with will limit/confine the multi-hop reasoning ability
%Focusing on single entity/relation/graph node only
Focusing only on a single token in the context and a single node in the graph is insufficient to enhance the multi-hop reasoning abilities of the dialogue system.
%the KG embeddings for Triple 1 and 2 are taken from the database gained by TransE and go through the multi-head attention mechanism. For the tokens "Great" in dialogue history, Triple 1 gains the higher context-knowledge cross attention score than Triple 2, because Triple 1 is more relevant with "Great". The output of the attention is treated as global knowledge grounding and will fuse into the token's textual embedding.}
In addition to local grounding, we further propose global knowledge grounding which enriches the semantics considering the entire sub-graph $\mathcal{G}_H$ and hence offers the model a comprehensive view of the background knowledge.

Following our observation, we adopt the attention mechanism~\cite{vaswani2017attention} to draw global dependencies between the dialogue history $H$ and a memory bank storing the representations of all knowledge triples in $\mathcal{G}_H$.
Let $T_H$ be the collection of all triples in $\mathcal{G}_H$. The memory bank stores  $|T_H|$ embedding vectors where the $i$-th vector $\vec{v}_i$ corresponds to the KG embedding of the $i$-th relation triple $T_i=\left \langle \texttt{[SBJ],[PRE],[OBJ]} \right \rangle$:
\begin{equation}
\vec{v}_i=M([SBJ])\oplus M([PRE])\oplus M([OBJ])
\end{equation}
where $\oplus$ is the concatenation operator for vectors.
%As shown in Figure \ref{fig:system}, we use TransE to gain the knowledge embedding $\mathbf{EMB_T}$ of triple $T$ and project the embedding for all triples to fed into the multi-head attention, as follows:
We gather all vectors and further project them to a global knowledge embedding space by:
\begin{equation}
\setlength\abovedisplayskip{3pt}%shrink space
\setlength\belowdisplayskip{3pt}
\begin{split}
K_H = W_{proj}\cdot[\vec{v}_1\oplus\cdots\oplus\vec{v}_{|T_H|}]
\end{split}
% \vspace{-0.5em}
\end{equation}
where $W_{proj}$ is a learnable projection matrix.

% 2*3*768 -> 2*768
% Here, $W_{proj}$ is a $3 \times 1$ learnable projection matrix, and $K_H$ is a $num\_triples \times embed\_dim$ resulting matrix for $\mathcal{G}_H$ that stores all knowledge for global grounding}, where $num\_triples$ is the number of triples in the memory bank and $embed\_dim$ is the dimension of the knowledge embedding.

%According to the following formula, the query is the textural embedding, key and value are both projected knowledge embedding for entire sub-graph. 
Based on the formulation of $K_H$, we compute how much attention the current token $w$ in $H$ pays to each relation triple according to the semantic relevance and obtain the~\emph{globally grounded} token embedding $\vec{w}_{global}$ as follows:
\begin{equation}
\setlength\abovedisplayskip{3pt}%shrink space
\setlength\belowdisplayskip{3pt}
\resizebox{1\hsize}{!}{$
\vec{w}_{global}= \left\{\begin{matrix}
\text{softmax}(\frac{\vec{w}\cdot K_H^T}{\sqrt{dim(\vec{w})}})\cdot K_H & \texttt{substr}(w, \mathcal{E}_H),
\\ & \texttt{substr}(w, \mathcal{R}_H)
\\
\vec{0} & otherwise
\end{matrix}\right.
$}
\end{equation}

As in Figure~\ref{fig:system} and Figure~\ref{fig:encoder}, for \texttt{``X-Men 2''} in the dialogue history, Triple 1 should have more influence on the tokens' representation than Triple 2 due to its higher relevance.
%the higher context-knowledge cross-attention score than Triple 2, because Triple 1 is more relevant with "Great".

Finally, the encoder of \textbf{RHO} sums the vanilla token embedding $\vec{w}$, the locally grounded embedding $\vec{w}_{local}$, and the globally grounded embedding $\vec{w}_{global}$ as:
$\tilde{w}= \vec{w}+\vec{w}_{local}+\vec{w}_{global}$.
During training, while $\vec{w}$ is rapidly updated via back propagation, $\vec{w}_{local}$ and $\vec{w}_{global}$ are relatively fixed with few parameters trainable (e.g.,~$W_{proj}$).

% \subsection{Training loss}
\subsection{Response Re-ranking}
\label{sec:re-rank}
With the above approaches, our knowledge-grounded model generates $N$ candidate responses by beam search. Yet, the grounding process mainly applies to the embedding level, lacking output constraints.
To enhance our \textbf{RHO}'s ability to reduce hallucination, we extend KG-CRUSE~\cite{sarkar2022kg} and train a conversational reasoning model $\phi$ for response re-ranking, with emphasis on the KG.
If the generated response can be reasoned backward to the source, we can assume it is faithful.
%This model explains/extracts the dialogue flow by strategically traversing over $\mathcal{G}_H$ conditioned on the dialogue history $H$ and a possible response $R$ and predicting the conditional probability of a random walk on $\mathcal{G}_H$.
%This reasoning model $\phi$ can predict the probabilities of a KG walk path composed of sequential knowledge triples for the given dialogue context which is the concatenation of dialogue history and response.

In our approach, we obtain the semantic embeddings of the dialogue history $H$ and a possible response $R$ via a contextual sentence encoder, i.e., Sentence-BERT~\citep{reimers2019sentence}.
The model $\phi$ is an LSTM-based decoder that learns the probability $p_{t,\phi}$ of an action $\vec{a}_t$ given the state $\vec{s}_t$ at step $t$. 
Here, the action refers to a walking step on the graph $\mathcal{G}_H$, represented as $\vec{a}_t$, which is the concatenation of the relation and entity embeddings derived from the KG, together with their semantic embeddings based on Sentence-BERT, i.e.,
\footnote{We have also investigated the impact of the KG embeddings ($\vec{e}_\mathcal{G}$ and $\vec{r}_\mathcal{G}$) for action modeling in Appendix~\ref{appendix sec: re-rank performance} by comparing the performance of the re-ranker under two settings: i) $\vec{a}_t = (\vec{e}_\mathcal{G} + \vec{e}_\mathcal{S})\oplus(\vec{r}_\mathcal{G} + \vec{r}_\mathcal{S})$ and ii) $\vec{a}_t = \vec{e}_\mathcal{S}\oplus \vec{r}_\mathcal{S}$ (the vanilla model in KG-CRUSE).}
%by comparing the performance of models with and without merging the KG embedding.
% which is the vanilla re-ranker in KG-CRUSE~\cite{sarkar2022kg}
\begin{equation}
\setlength\abovedisplayskip{3pt}%shrink space
\setlength\belowdisplayskip{3pt}
\vec{a}_t = (\vec{e}_\mathcal{G} + \vec{e}_\mathcal{S})\oplus(\vec{r}_\mathcal{G} + \vec{r}_\mathcal{S})
\end{equation}
where $\vec{e}_\mathcal{S}$ and $\vec{r}_\mathcal{S}$ are the semantic sentence embeddings of an entity $e$ and a relation predicate $r$, respectively. 
The state $\vec{s}_t$ contains the representations of the dialogue history, together with entities and relations already traversed by $\phi$ (action history). It is defined as a tuple $(H, (\vec{a}_1, \vec{a}_2, \cdots, \vec{a}_{t\mbox{-}1}))$.
Hence, the model $\phi$ explicitly models the process of a traversal upon $\mathcal{G}_H$ conditioned on the dialogue history $H$ and a possible response $R$. 
During training, each action $\vec{a}_t$ made by $\phi$ is combined into a path, and the target path is the given context-related sub-graph $\mathcal{G}_H$.
%And the sequential information in $\mathbf{s}_t$ is modelled via LSTM.

%While for entities and relations in KG, in addition to the semantic sentence embedding, we also utilise the KG embedding ($\vec{e}_\mathcal{G}$ and $\vec{r}_\mathcal{G}$) learned by TransE in the pre-processing step (Section~\ref{sec:local})~\footnote{We investigate the impact of the KG embedding in Appendix~\ref{appendix sec: re-rank performance} by comparing the performance of models with and without merging the KG embedding.}. 

% $\mathbf{a}_t\in \mathbf{A}_t$
% And $\mathbf{A}_t$ is the set of all possible actions available including the outgoing end/outlet of $\mathbf{a}_{t-1}$ from $\mathcal{G}_H$. 

% \begin{equation}
% % \setlength\abovedisplayskip{3pt}%shrink space
% % \setlength\belowdisplayskip{3pt}
% p_{t,\phi}(\mathbf{a_t}|\mathbf{s_t},\mathbf{A_t}) = \frac{exp(\mathbf{a}_t\cdot \mathbf{x}_t)}{\sum _{a_i\in A_t}exp(\mathbf{a}_i\cdot \mathbf{x}_t)}
% % \vspace{-0.5em}
% \end{equation}
% where $\mathbf{x}_t$ is the final context vector 

After our encoder-decoder model generates $N$ candidate responses $\{ R_1,\cdots, R_N\}$, we select the best response $R^*$ with the highest probability $\mathbf{p_\phi} = \prod_t p_{t,\phi}$ over all the generated responses, i.e.,
\begin{equation}
\setlength\belowdisplayskip{3pt}
R^* = \mathop{\arg\max}\limits_{n\in \{1,\cdots,N\}}\mathbf{p}_{\phi}(\mathbf{A}=\mathcal{G_H}|H,R_n)
% \vspace{-0.5em}
\end{equation}
where $\mathbf{A}$ is a collection of actions $\vec{a}_t$ (i.e. knowledge path) that $\phi$ has already traversed conditioned on the dialogue history $H$ and each response $R_n$. 

For a more intuitive understanding, refer to the example in Figure~\ref{fig:system} where the model generates three candidate responses: A, B, and C. It selects Response C as the final output with the traversal path \texttt{``X-Men 2, directed\underline{ }by, Bryan Singer, $\sim$directed\underline{ }by, Superman Returns''}.
% \textit{``Great\underline{ }Expectations, written\underline{ }by, Charles\underline{ }Dickens, written\underline{ }by$^{-1}$, A\underline{ }Christmas\underline{ }Carol''}
\footnote{$\sim$directed\underline{ }by refers to the opposite direction of the relation directed\underline{ }by.}
As seen, there is a higher matching degree between the sub-graph in Figure~\ref{fig:system} and the Response C, compared to other candidate responses (i.e.,~A and B).
%For each response, input the dialogue history and response into re-ranker which then obtains the probability of extract the knowledge path "<Great Expectations, written by, Charles Dickens>, <A Christmas Carol, written by, Charles Dickens>". Finally, the re-ranker selects response C because the conversation flow with it follows the path.
  
\section{Experiments}
\subsection{Dataset}
% authority classic and why only this?
\textbf{OpenDialKG}~\citep{moon2019opendialkg} contains open-ended dialogues between two speakers, initiated by talking about a given entity and grounded on the relevant facts from a structured KG. Thus, the sequential turn-based dialogues can be regarded as traversing the paths in the KG. 
% DuConv~\citep{wu2019proactive} also grounded dialgoue on KG paths. However, it contains Chinese Proactive conversations where one speaker acts as a conversation leader and the other acts as the follower.
To our knowledge, OpenDialKG is currently the only publicly available corpus for English open-ended dialogues with KG path annotations~\cite{yu2022survey,ni2022recent}, and previous works~\citep{dziri-etal-2021-neural,zhou2021earl} evaluate their effectiveness on this corpus. Hence, we also conduct our experiments on OpenDialKG.
%and those triples represent either 1-hop or 2-hop paths from the initial entity.
Consistent with previous works~\citep{dziri-etal-2021-neural,liu2019knowledge,zhou2021earl}, we filter OpenDialKG by keeping only the dialogue samples that are annotated with a KG path. The dataset is divided into training, validation, and testing sets in the ratio of 8:1:1.

\subsection{Baselines}
\label{sec:baseline}
The following strong baselines are employed to show the efficiency of our method.
We fine-tune pre-trained language models \textbf{GPT2}~\cite{radford2019language} and \textbf{BART}~\citep{lewis2020bart} on our task.
% As we mentioned in Section~\ref{sec: related}, 
NPH~\citep{dziri-etal-2021-neural} refines the generated responses by retrieved entities from the KG. 
 To our knowledge, the integration of GPT2 and NPH, called \textbf{GPT2+NPH}, reaches the SOTA performance on OpenDialKG.
Since this post-processing technique is agnostic to the generation model, we apply it to BART, named~\textbf{BART+NPH} as our baseline.
In addition,
\textbf{EARL}~\cite{zhou2021earl} utilizes external KGs for conversation generation without parameterizing specific entity representations.
% EARL~\cite{zhou2021earl} masks entities in conversation history and utilizes the relational structure of the KG to learn the category representations for entities.
%so that the entity-agnostic representation can generalize to unseen entities.
\textbf{KG-BART}~\citep{liu2021kg}, a KG-augmented pre-trained language generation model based on BART, introduces the information of the relations among concepts for generative commonsense reasoning. 
We are the first to adapt this model to the KGD generation.
% and first/originally 
Furthermore, we explore \textbf{ChatGPT} on this task in Appendix~\ref{appendix sec: implement}.
Please refer to it for the details of baseline implementations.

\begin{table*}[!t]
\centering
\resizebox{0.95\linewidth}{!}{
\begin{tabular}{lcccccccc}
\toprule
\multicolumn{1}{c}{\multirow{2}{*}{\textbf{Model}}} & \multicolumn{1}{c}{\multirow{2}{*}{\textbf{BLEU4 $\uparrow$}}} & \multicolumn{1}{c}{\multirow{2}{*}{\textbf{ROUGE-L $\uparrow$}}} & \multicolumn{1}{c}{\multirow{2}{*}{\textbf{FeQA $\uparrow$}}} & \multicolumn{2}{c}{\textbf{QuestEval $\uparrow$}} & \multicolumn{3}{c}{\textbf{Entity Coverage (\%) $\uparrow$}}                  \\
\multicolumn{1}{c}{}                       & \multicolumn{1}{c}{}    & \multicolumn{1}{c}{} & \multicolumn{1}{c}{}                                  & \multicolumn{1}{c}{\textbf{RD}} & \multicolumn{1}{c}{\textbf{RF}} & \multicolumn{1}{c}{\textbf{Pre.}} & \multicolumn{1}{c}{\textbf{Recall}} & \textbf{F1} \\ \hline
EARL~\cite{zhou2021earl} & 7.97	& 23.61 & 39.93 & 37.88&35.59 & 86.61&45.17&64.44\\
% GPT2~\cite{radford2019language} & 10.27 & 29.59 & 39.60* & 46.86&42.07 & 91.62&33.26&52.30 \\
% GPT2+NPH~\cite{dziri-etal-2021-neural} & 10.41 & 29.93 & 40.83* & 47.45&42.45 & 95.61&33.39&53.96  \\
GPT2~\cite{radford2019language} & 10.27 & 29.59 & 39.60/26.54$^\dagger$ & 46.86&42.07 & 91.62&33.26&52.30 \\
GPT2+NPH~\cite{dziri-etal-2021-neural} & 10.41 & 29.93 & 40.83/28.98$^\dagger$ & 47.45&42.45 & 95.61&33.39&53.96  \\
% GPT2~\cite{radford2019language} & 10.27/11.10* & 29.59/30.00* & 39.60/26.54* & 46.86&42.07 & 91.62&33.26&52.30 \\
% GPT2+NPH~\cite{dziri-etal-2021-neural} & 10.41/11.29* & 29.93/31.00* & 40.83/28.98* & 47.45&42.45 & 95.61&33.39&53.96  \\
BART~\cite{lewis2020bart} & 14.45 & 33.33 & 39.00 & 46.97&42.75 & 96.99&44.96&62.87  \\
BART+NPH & 15.53 & 34.99 & 42.41 & 47.94&43.56 & 96.44&44.12&65.98 \\
KG-BART~\cite{liu2021kg} & 13.72 & 33.31 & 41.87 & 45.55&42.86 & 97.68&45.63&64.58  \\
\hline
\textbf{RHO} (LKG) & 19.89 & \textbf{39.95} & 43.04 & 48.91&44.37 & 97.38&45.57&67.77  \\%only entity kg embedding local w/o. RR and GKG
\textbf{RHO} (GKG) & \textbf{20.77} & 39.54 & 40.65 & 48.41&43.84 & 97.20&45.63&67.40  \\%only memory bank global w/o. RR and LKG
\textbf{RHO} (LKG+GKG) & 20.63 & 39.51 & 45.96 & 50.35&46.03 & 98.26&50.74&71.47 \\%w/o. RR
\textbf{RHO} (Full Implementation) & 19.11 & 38.45 & \textbf{47.99} & \textbf{50.58}&\textbf{46.41} & \textbf{98.53}&\textbf{51.77}&\textbf{72.29} \\
\bottomrule
\end{tabular}
}
% \quad\quad
\caption{Automatic evaluation results for \textbf{RHO} and baselines, where ``RD'', ``RF'', and ``Pre.'' refer to reference-dependent, reference-free mode, and Precision, respectively.
The results of the ablation study are shown in the last four rows.
``LKG'', ``GKG'' and ``RR'' refers to local knowledge grounding, global knowledge grounding and response re-ranking, respectively.
``Full Implementation'' means that we implement all three components, i.e., LKG+GKG+RR.
$\dagger$ The FeQA scores we calculate (former) are higher than those reported in~\citet{dziri-etal-2021-neural} (latter).
% *The FeQA scores we calculate are higher than those reported in~\citet{dziri-etal-2021-neural} (26.54 for GPT2 and 28.98 for GPT2+NPH).
}%,madotto2021few
\label{tab:auto_result}
\vspace{-1em}
\end{table*}

\subsection{Evaluation Protocols}
% Experimental settings ??? and metrics
\subsubsection{Automatic Evaluation}
To evaluate the generation quality, we use the classical word overlap-based NLG metrics: \textbf{BLEU} and \textbf{ROUGE-L}~\citep{lin2004rouge}.
Due to the possible presence of hallucinations in the dataset (especially extrinsic ones) the metrics based on the n-gram overlap between the golden answer and generated texts are not sufficient~\cite{ji2022survey}. 
% we should not rely deeply on the metrics based on the grams overlap between golden answer and generated text
Therefore, we also use source-dependent metrics, i.e.,~\textbf{FeQA}~\citep{durmus-etal-2020-feqa},~\textbf{QuestEval}~\citep{scialom2021questeval}, and \textbf{Entity Coverage}, to estimate the hallucination degree.
\textbf{FeQA} and \textbf{QuestEval} are both question-answering (QA)-based metrics for evaluating the faithfulness of the output in generation tasks~\footnote{Please refer to Appendix~\ref{appendix sec: automatic metric} for details.}. 
\textbf{QuestEval} has two modes: reference-dependent mode assesses a text with one or multiple ground-truth references; and reference-free mode assesses a text without any references.
% depends both on input source and golden answers.
In addition to the metrics used in previous works~\citep{dziri-etal-2021-neural, zhou2021earl}, we assume that entities in generated responses should be covered by those in the given knowledge triples and dialogue history. The higher \textbf{Entity Coverage} is, the lower hallucination degree can be to some extent. Specifically, we utilize a named entity recognition (NER) model to extract named entities in generated responses and the dialogue history.
We compute \textbf{Entity Precision, Recall and F1} scores between generated entities and entities in KG and dialogue history to evaluate the faithfulness of generated responses.

% BLEU is a n-gram based model which usually used in natural language processing task to evaluate the distance between generated sentence and ground truth. It ranges from 0 to 1, the bigger, the better. 
% ROUGE is a evaluating metric based on the longest common subsequence. It can used to compare similarity between generated sentence and ground truth. 

\subsubsection{Human Evaluation}
To further assess the quality of generated responses from different systems, we conduct human evaluations using Amazon Mechanical Turk\footnote{\url{https://www.mturk.com/}\label{Turk}}.
For hallucination level assessment, we first ask annotators to identify whether each response is \textbf{Faithful}, or \textbf{Hallucinated} given the dialogue history and KG triples.
% , or \textbf{general}.
The judgment criteria are as described in \S~\ref{sec:introduction}.
``Faithful'' means that the response is supported by the knowledge triples and dialogue context, while ``hallucinated'' means that the response contradicts or cannot be verified by the source.
% , while "general" means the response does not include any facts (e.g., I don’t know about it).
If the response is hallucinated, we further ask annotators to identify whether the hallucination is \textbf{Extrinsic}, \textbf{Intrinsic} or \textbf{Both}~\citep{dziri-etal-2021-neural}.
% The settings of human hallucination assessment references~\citet{dziri-etal-2021-neural}. 
% can be further found in
% The setting of human evaluation for hallucination references~\citet{dziri-etal-2021-neural}.
% validate the effectiveness of our method

We also conduct an A/B test of our framework against the baselines GPT2+NPH and BART+NPH to evaluate generated responses on \textbf{Fluency}~\citep{ji2022vscript,dathathri2019plug}. 
The annotators are asked whether the writing is smooth and grammatically correct and given four choices: \textbf{Neither}, \textbf{Both}, \textbf{Sample A}, or \textbf{Sample B}. 
Please refer to Appendix~\ref{appendix sec: human eval} for details.

\section{Results and Analysis}
\subsection{Overall Evaluation Results}
\label{sec:results}
\noindent\textbf{Automatic Evaluation.}
% Overall Results
The first eight rows of Table~\ref{tab:auto_result} shows the experimental results on automatic metrics over the OpenDialKG test set.  Our model outperforms all baselines on both classic overlap metrics and hallucination metrics, indicating the high quality of the generated utterances.
Specifically, compared to SOTA (GPT2+NPH), \textbf{RHO} gives a significant rise of 17.54\% in FeQA, 9.33\% in QuestEval (RF), and 33.97\% in Entity Coverage (F1).
\textbf{RHO} also achieves better results compared to the stronger baseline BART+NPH.
The results indicate the faithfulness of KGD systems can be improved by knowledge grounding and re-ranking techniques.
% We note that results with BART as the backbone are better than GPT2.
% GPT2 is a decoder model in an auto-regressive manner, while BART contains a bidirectional encoder which makes it more advantageous in global dependency and long-term understanding~\citep{wang-etal-2022-chae,shakeri-etal-2020-end,liu2021kg}.
% % ,lewis2020bart

\begin{table}[!t]
\centering
\resizebox{0.9\linewidth}{!}{
\begin{tabular}{lccccc}
\toprule
\multicolumn{1}{c}{\multirow{2}{*}{\textbf{Model}}} & \multicolumn{1}{c}{\multirow{2}{*}{\textbf{Faith. (\%) $\uparrow$}}} & \multicolumn{3}{c}{\textbf{Hallucination (\%) $\downarrow$}} \\
\multicolumn{1}{c}{} & \multicolumn{1}{c}{} & \multicolumn{1}{c}{\textbf{In.}} & \multicolumn{1}{c}{\textbf{Ex.}} & \multicolumn{1}{c}{\textbf{Both}}   \\ \hline
GPT2+NPH & 72.67 & 8.67 & 18.00 & 0.67 \\
BART+NPH & 75.00 & 9.33 & 15.33 & \textbf{0.33} \\
\textbf{RHO} w/o RR & 80.67 & \textbf{7.67} & 10.67 & 1.00 \\
\textbf{RHO} & \textbf{81.67} &  \textbf{7.67} & \textbf{10.00} & 0.67 \\
\hline
\end{tabular}
}
% \bottomrule
\caption{Human evaluation results for hallucination degree, where ``Faith.'', ``In.'', and ``Ex.'' refers to faithfulness, intrinsic, and extrinsic hallucination, respectively.}
\vspace{-0.5em}
\label{tab:human_eval_hallucination}
\end{table}

\begin{table}[!t]
\centering
\resizebox{\linewidth}{!}{ 
\begin{tabular}{@{}ccccc@{}}
\toprule
\diagbox[width=0.4\linewidth]{\textbf{Model}}{\textbf{Fluency}} 
% \textbf{Model} 
& \textbf{Win (\%)$\uparrow$} & \textbf{Lose (\%)$\downarrow$}& \textbf{Tie (\%)}\\ \midrule
\textbf{RHO} w/o RR vs. GPT2+NPH & 37.33 & 20.67 & 42.00 \\ 
\textbf{RHO} w/o RR vs. BART+NPH & 24.67 & 18.67 & 56.67 \\
\textbf{RHO} vs. GPT2+NPH & 32.33 & 16.00 & 51.67 \\ 
\textbf{RHO} vs. BART+NPH & 17.00 & 12.67 & 70.33 \\ 
\bottomrule
\end{tabular}
}
\caption{Human evaluation results for fluency.}
\vspace{-1em}
\label{tab:human_eval_fluency}
\end{table}

\begin{table*}[!t]
\centering
\resizebox{1\linewidth}{!}{
\begin{tabular}{lllll}
\toprule
\multicolumn{1}{c}{\textbf{KG}} & \multicolumn{1}{c}{\textbf{History}} & \multicolumn{1}{c}{\textbf{Golden}} & \multicolumn{1}{c}{\textbf{GPT2+NPH}} & \multicolumn{1}{c}{\textbf{RHO}} \\ \hline
\begin{tabularx}{0.15\textwidth}{X}<The Ref, \\ starred actors, \\ Judy Davis> \end{tabularx} &
\begin{tabularx}{0.35\textwidth}{X} User: Who else was in that movie? \\
Assistant: Aside from Woody Allen, Judy Davis and Greta Gerwig also have star credits. \\
User: What else do I know Judy Davis from? \end{tabularx} &
\begin{tabularx}{0.17\textwidth}{X} Assistant: She also starred in "The Ref" and "\textbf{My Brilliant Career}." \end{tabularx} &   
\begin{tabularx}{0.17\textwidth}{X} Assistant: She starred in \textbf{The Referendum} and \textbf{The Golden Compass}. \end{tabularx} &     
\begin{tabularx}{0.17\textwidth}{X}Assistant: Judy Davis also starred in The Ref. Have you seen this movie?\end{tabularx} \\ 
\bottomrule
\end{tabular}
}
\caption{An example response generated by GPT2+NPH and our model. \texttt{``My Brilliant Career''} in the golden answer, \texttt{``The Referendum''} and \texttt{``The Golden Compass''} from NPH are extrinsic hallucinations since there is no clue in external knowledge and dialogue history. }
\vspace{-1em}
\label{tab:positive case}
\end{table*}

\noindent\textbf{Human Evaluation.}
% ~\footnote{The result is statistically significant with p < 0.05.\label{pvalue}}
As in Table \ref{tab:human_eval_hallucination}, the faithfulness of \textbf{RHO} is higher than GPT2+NPH and BART+NPH. 
The results are statistically significant with p-value$<$0.05. 
Specifically, we see a 12.38\% increase in faithfulness compared to SOTA (GPT2+NPH).
As shown in Table~\ref{tab:positive case}, the information \textit{``Judy Davis starred in My Brilliant Career''} in golden answer, is not supported by the input, although it is factual according to the world knowledge.
The baseline model hallucinates unfactual information, i.e., \textit{``Judy Davis starred in both The Referendum and The Golden Compass''}, while our model is better grounded on the input source. 

At a more granular level, the extrinsic hallucination problem is more frequent than the intrinsic one in all models. This phenomenon is also observed in other works~\citep{dziri-etal-2021-neural,nan2021improving}.
% in dialogue~\cite{dziri-etal-2021-neural} and summarization~\cite{nan2021improving}. 
%The reason may be that the corpus itself has more extrinsic hallucination and the model is more likely to produce unverified rather than contradictory information with the limitation of source. 
Specifically, compared to SOTA (GPT2+NPH), \textbf{RHO} reduces extrinsic hallucination by 42.85\%. Compared to BART+NPH, \textbf{RHO} reduces intrinsic hallucination by 13.66\% and extrinsic hallucination by 46.74\%.
%fluency
According to the A/B test results for fluency in Table \ref{tab:human_eval_fluency}, \textbf{RHO} is slightly more fluent than SOTA methods.
Overall, human evaluation results are in line with automatic evaluation. \textbf{RHO} mitigates both intrinsic and extrinsic hallucination issues without sacrificing fluency.

\subsection{Ablation Study}
We conduct an ablation analysis to assess the contribution of each component of our method: Local Knowledge Grounding (LKG), Global Knowledge Grounding (GKG), and Response Re-ranking (RR). 
%  overlap-based
As shown in the last four rows of Table \ref{tab:auto_result}, fully-implemented \textbf{RHO} performs best in automatic hallucination metrics with a slight sacrifice of classical overlap metrics.
Specifically, compared to models equipped with only local/global knowledge grounding, the model equipped with both two (LKG+GKG) gains higher scores in FeQA, QuestEval, and Entity Coverage. 
The same trend is observed when comparing the fully-implemented model with the model without re-ranking (LKG+GKG) and please refer to Appendix~\ref{appendix sec: re-rank case} for an example.
The trade-off between hallucination metrics and the others is due to the fact that some reference responses in the dataset diverge from the sources~\citep{dziri2022origin}. Hence, improving responses' consistency with the source will inevitably reduce that with the references.
Overall, the above three mechanisms synergy to improve the generated responses' faithfulness without significant reduction in quality. 
% (need?) The local grounding is effective to take advantage of lexical and knowledge information; the global grounding is able to equip KGD systems with the multi-hop reasoning; and re-ranking effectively enforce the model to follow the knowledge traverses throughout KG for the whole conversation.

% % grounding
% Compared with models equipped with only local/global knowledge grounding, the model equipped with both two gains higher scores in FeQA and Entity Coverage. 
% This indicates that local and global knowledge grounding synergy to improve the generated responses' faithfulness without significant reduction in overall quality.
% Therefore, the local grounding is effective to take advantage of lexical and knowledge information
% and the global grounding is able to equip KGD systems with the multi-hop reasoning.
% % re-ranking
% Compared with the model without re-ranking, \textbf{RHO} gain higher scores in the hallucination metrics, while a little lower scores in others.
% Therefore, re-ranking effectively enforce the model to follow the knowledge traverses throughout KG for the whole conversation.
% Please refer to the example in Appendix~\ref{appendix sec: re-rank case}.

% \subsection{Sensitivity analysis}
% Performance on dev. w.r.t. hyper-parameters (a few figures)

\subsection{Error Analysis}
\label{sec:error analysis}
% case study and 
% Characterize some types of errors, percentage in samples, provide error cases, and guide lines for future improvement
Although \textbf{RHO} achieves better performance than the baselines in the KDG task, it still generates a few cases of failure. To gain more insight into our model, we present failure examples from \textbf{RHO} in Appendix~\ref{appendix sec: error analysis} and conduct an error analysis.

% There are extrinsic hallucinations both in golden answer and our response. 
As shown in Table~\ref{tab:human_eval_hallucination}, the extrinsic hallucination problem is more frequent. 
%In our generated responses, the incidence is about 10\% and 8\%, respectively.
%This phenomenon is also observed in other works in dialogue~\cite{dziri-etal-2021-neural} and other fields such as summarization~\cite{nan2021improving}. 
Because the corpus itself has more extrinsic hallucinations and the model is more likely to produce unverified rather than contradictory information with the limitation of the source. 
% The contradictions and errors in dataset make dialogue model confused and reason out wrong information such as reversed relationship.
% The NLG model is trained on noisy data, and the open-domain dialogue task innately value diversity in generated output~\cite{ji2022survey}.
To solve the problem, we suggest that the quality of the corpus used for model training, such as OpenDialKG, can be further improved by filtering out irrelevant and contradictory information.
%pre-possing data is intuitive method which finds irrelevant and contradictory information from the existing corpus and then filters it. 
Fact-checking can also be potentially used to find evidence from world knowledge and verify the generated responses~\cite{ji2022survey}.

In addition to intrinsic and extrinsic hallucination issues, we discover two other types of negative cases, namely \textbf{No Full Coverage}, and \textbf{Unnatural Connection}.
\textbf{No Full Coverage} refers to the situation where
the output does not cover the full answer as expressed by all triples mentioned in the given sub-graph.
%as No Full Coverage.
% Based on our analysis, about 10\% of the generated responses miss some information of the given knowledge.  
We believe that further research on the reasoning ability and interpretability of the model can help address this issue. 
%We assume that the knowledge path represents the flow and reasoning of the entire conversation, but 
\textbf{Unnatural Connection} denotes that a response is not connected to the dialogue history especially the last utterance naturally.
We find that in a few cases, the dialogue system delivers an irrelevant answer or poorly replies to the dialogue history.
%The dialogue system gives an irrelevant answer or reply to the dialogue history and the probability is about 16.00\%.
This issue occurs because our model focuses more on hallucination reduction and sometimes pays less attention to flexibility and diversity. 
In the future, we can explore the controllability of the grounding degree on knowledge and strike a dynamic balance between faithfulness and flexibility in response generation.

\section{Conclusion and Future Work}
In this work, we investigate the hallucination in KGD systems and propose the \textbf{RHO} framework to tackle this issue
 with three mechanisms: Local Knowledge Grounding, Global Knowledge Grounding, and Response Re-ranking. 
% % This model is equipped educe hallucination effectively
Our method is empirically proven effective in reducing hallucinations with automatic and human evaluation. 
% This is also consistent with the results of human evaluation.
We also conduct deep error analysis on the generated responses. 
% Future works include exploring better knowledge representation learning techniques, applying knowledge grounding to model decoding, and obtaining a better balance between response faithfulness and flexibility.
For future work, the re-ranking module can be combined with foundation models such as ChatGPT to reduce manual labor in the human feedback process. We also call for research to strike a better balance between response faithfulness and diversity.

%suggest future improvement directions for each category.

\section*{Limitations}
\label{sec:Limitations}
% This work does not gain the graph embedding by leveraging the emerging deep neural networks such as GAT~\cite{velivckovic2017graph}, GCN~\cite{kipf2016semi}.

%不要放technique detail的问题 不要cite其他paper
The deep neural networks in \textbf{RHO} uses feature extraction and vectorization to represent the texts.
The model only detects the statistical regularities and quantitative relationships among the variables but can not see qualitative relationships, such as causality, hierarchy, and other abstractions~\citep{tsimenidis2020limitations}.
Although we leverage the response re-ranking technique, which improves the explainability of \textbf{RHO}, the neural networks are undoubtedly still ``black boxes'' to humans. 
Therefore, the faithfulness of generated responses can not be fully guaranteed.

\section*{Ethical Considerations}
\label{sec:ethical}
% Authors are granted extra space after the 8th page for an ethics/broader impact statement

% Intended use. If the technology is functioning as intended, who benefits? Who might be harmed, and how?
% Failure modes. What are the failure modes, and in case of failure, who might be harmed and how?
% Biases. What are the biases included in the dataset and trained model, and how might they contribute to those failure modes? (On bias in NLP, see Blodgett et al 2020.)
% Misuse potential. What kinds of potential misuse is there for the technology and what harms might ensue? What could be done to prevent such misuse/what should regulators know about this technology?
% Potential harm to vulnerable populations. Are any of the possible harms you’ve identified likely to fall disproportionately on populations that already experience marginalization or are otherwise vulnerable?

%As we mentioned in Section~\ref{sec:error analysis}, there are still complex samples that \textbf{RHO} cannot handle. These failure cases can be misleading and detrimental to user experience.

In addition to the hallucination problem, another critical challenge, the offensive language, is also introduced with the evolutionary progress toward building reliable dialogue systems. 
The data-driven models are susceptible to delivering offensive responses while mimicking human conversations~\cite{xu2020recipes}. 
% For example, the Microsoft chatbot Tay had to be dragged down from Twitter less than 24 hours after its launch due to sexist and racist tweets generated by the system. 
It has been shown that racial and gender biases are encoded in the PLMs~\cite{blodgett-etal-2020-language}, and these biases are present in the training corpus.
Since \textbf{RHO} leverages PLMs and the training corpus, it is possible to generate offensive languages.
We suggest that in real-world dialogue systems, it is necessary to employ some post-processing steps to alleviate this problem when it is deployed online.  
% but our paper does not improve the control over it.

% ~\cite{henderson2018ethical}

\section*{Acknowledgement}
\label{sec:Acknowledgement}
This work has been supported by the China NSFC Project (No. NSFC21EG14), SAAIR Project (No. Z1286), and HKJCCT21EG01 (RG192).

\bibliography{custom}
\bibliographystyle{acl_natbib}

\appendix

\clearpage
\section{Implementation Details}
\label{appendix sec: implement}
\paragraph{RHO}
The maximum dialogue history length is set to 3 utterances. This setting is also held constant in baselines. 
Our method is implemented using the Huggingface Transformers library\footnote{\url{https://github.com/huggingface}}. 
We load the pre-trained BART-base model and train \textbf{RHO} with the following settings and hyper-parameters: the batch size 16, the learning rate 3e-5, and the AdamW optimizer with a linear scheduler. We generate multiple candidate responses using beam search (with the number of beams B=4).
Our model is trained on one NVIDIA Geforce RTX 3090 GPU. It takes approximately 3 hours to train.

\paragraph{BART}
We fine-tune a BART-base model with the following settings and hyper-parameters: the batch size 16, the learning rate 3e-5, and the AdamW optimizer with a linear scheduler.
We also generate responses using beam search (B=4).

\paragraph{Modified KG-BART}
We modify the code from the official library\footnote{\url{https://github.com/yeliu918/KG-BART}} to fit our KGD task. We load the pre-trained BART-base model and train the modified KG-BART with the default hyper-parameters: the batch size 16, the learning rate 1e-5, and the AdamW optimizer with a linear scheduler.
We also generate responses using beam search (B=4).

\paragraph{GPT2}
We fine-tune GPT2-small with the following settings and hyperparameters: the batch size 16, the learning rate 6.25e-5, and the AdamW optimizer with a linear decay scheduler. We also generate responses using beam search (B=4).
More details of hyper-parameters can be found in~\citet{dziri-etal-2021-neural} where GPT2 is regarded as a strong baseline for the same task.

\paragraph{NPH}
We implement NPH using the code from the official library\footnote{\url{https://github.com/nouhadziri/Neural-Path-Hunter}} with their default hyper-parameters. We also utilize the provided graph embeddings and the hallucination critic model.

\paragraph{EARL}
We obtain the best-generated responses from the authors of~\citet{zhou2021earl} and evaluate the quality of the responses via our metrics.

\paragraph{ChatGPT}
We randomly select 50 samples from the test set of OpenDialKG. 
Prompt engineering is needed when applying ChatGPT to our task.
At first, ChatGPT refuses to generate a response because it is too cautious to express opinions or feelings.
For example, we input \texttt{``Given the knowledge: We Rode in Trucks, Composer, Luke Bryan\textbackslash n \textbackslash n User: I like Luke Bryan's music. What do you think about him?\textbackslash n \textbackslash n Please generate the next turn.''}
The output is \texttt{``As an AI, I do not have personal opinions or feelings. However, I can provide information about Luke Bryan's music and career...''}

Therefore, we try different prompts, and the successful one we adopt is \texttt{``Can we try dialogue generation? I will give you turns, and you can generate the next turn, but only one.\textbackslash n \textbackslash n You can also consider the knowledge of "We Rode in Trucks, Composer, Luke Bryan" for your reference in the dialogue.\textbackslash n \textbackslash n User: I like Luke Bryan's music. What do you think about him?''}
The output is \texttt{``Assistant: I think Luke Bryan is a talented musician and songwriter. His hit songs like \textbf{"Country Girl (Shake It for Me)"} and "We Rode in Trucks" showcase his ability to connect with audiences through his music.''}

% According to human judgment, the responses from ChatGPT are of high quality. 
% However, the automatic evaluation results in Table~\ref{tab:chatgpt} are relatively low. 
After human observation and analysis, we find the following phenomena:
The generated responses tend to be long, sometimes even generating multiple turns, even if we ask it to generate only one. The length distinction from the golden answer. % leads to low scores in reference-dependent metrics
In addition, ChatGPT refers to both the given knowledge and the parametrized background knowledge injected during pre-training. There are lots of extrinsic hallucinations that cannot be verified and supported by the input source. 
For example, \texttt{``Country Girl (Shake It for Me)''} in the previous paragraph. 
% Most of them are factual, according to manual checks with Wikipedia.
More exploration is needed on detecting and mitigating hallucination in ChatGPT on KGD.

% \begin{table*}[!t]
% \centering
% \resizebox{0.95\linewidth}{!}{
% \begin{tabular}{lcccccccc}
% \toprule
% \multicolumn{1}{c}{\multirow{2}{*}{\textbf{Model}}} & \multicolumn{1}{c}{\multirow{2}{*}{\textbf{BLEU4 $\uparrow$}}} & \multicolumn{1}{c}{\multirow{2}{*}{\textbf{ROUGE-L $\uparrow$}}} & \multicolumn{1}{c}{\multirow{2}{*}{\textbf{FeQA $\uparrow$}}} & \multicolumn{2}{c}{\textbf{QuestEval $\uparrow$}} & \multicolumn{3}{c}{\textbf{Entity Coverage (\%) $\uparrow$}}                  \\
% \multicolumn{1}{c}{}                       & \multicolumn{1}{c}{}    & \multicolumn{1}{c}{} & \multicolumn{1}{c}{}                                  & \multicolumn{1}{c}{\textbf{RD}} & \multicolumn{1}{c}{\textbf{RF}} & \multicolumn{1}{c}{\textbf{Pre.}} & \multicolumn{1}{c}{\textbf{Recall}} & \textbf{F1} \\ \hline
% ChatGPT & 4.05	& 18.62 & 15.03 & 44.25&43.01 & 67.63&66.15&62.72 \\
% \bottomrule
% \end{tabular}
% }
% % \quad\quad
% \caption{Automatic evaluation results for baseline ChatGPT on 50 test samples, where ``RD'', ``RF'', and ``Pre.'' refer to reference-dependent mode, reference-free mode, and Precision, respectively.}
% \label{tab:chatgpt}
% \vspace{-1em}
% \end{table*}

\section{Evaluation}

\subsection{Automatic Metrics}
\label{appendix sec: automatic metric}
\paragraph{FeQA}
\label{appendix sec: feqa}
FeQA, a QA-based metric for evaluating the faithfulness of the generated output, has been applied in summary~\cite{chen2021improving} and dialogue~\cite{dziri-etal-2021-neural} tasks.
As a reference-free metric, it takes the source (such as a document) and the corresponding output to be evaluated (such as a summary) as input. 
Given the source, a question generation (QG) model generates a question based on the source and then a QA model generates an answer A. The QA model generates another answer B based on the question and output to be evaluated.
The average F1 score between answers A and B reflects the hallucination level of the output.
Following the setting in~\citep{dziri-etal-2021-neural}, we concatenate all the knowledge triples in $\mathcal{G_H}$ with the dialogue history $H$ as the source, and the generated response is the output.
We calculate the FeQA score using the code and models from the official library\footnote{\url{https://github.com/esdurmus/feqa}} with their default hyper-parameters.

% The FeQA scores we calculate in Table~\ref{tab:auto_result} are higher than the ones reported in~\citet{dziri-etal-2021-neural}. For GPT2, we measure 39.60 compared to their 26.54; and for GPT2+NPH we measure 40.83 compared to their 28.98.
% % Although it is not stated in the paper~\cite{madotto2021few}, the beam size for FeQA their in source code~\footnote{\url{https://github.com/andreamad8/FSB/blob/2931c6c97aeda0299472e06802880a6a1ec81d14/metric/feqa.py}} is 1. The beam size we use is the default value of 10, and we think the gap comes from this.

\paragraph{QuestEval}
QuestEval~\citep{scialom2021questeval} is also a QA-based metric that has reference-dependent and reference-free modes.
The reference-dependent mode depends both on the input source and golden answers, while the reference-free mode does not require any ground-truth references. The input source is built in the same way as in FeQA.
We calculate the QuestEval score using the code and models from the official library\footnote{\url{https://github.com/ThomasScialom/QuestEval}} with their default hyper-parameters.

\subsection{Human Evaluation}
\label{appendix sec: human eval}
 
\begin{figure*}[!ht]
 \centering
 \includegraphics[width=1\linewidth]{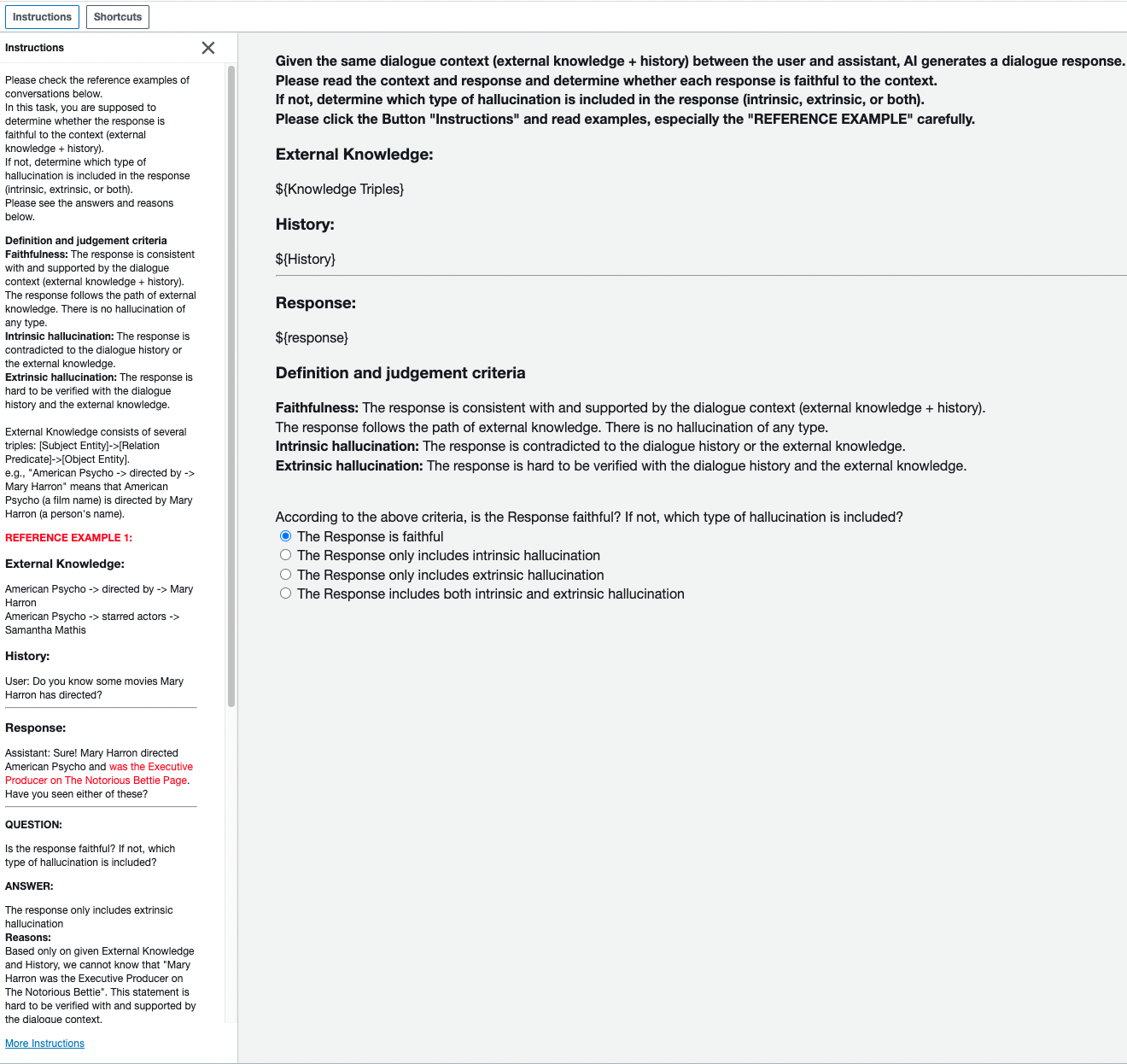}

  \caption{The UI for human evaluation on hallucination.}
  \label{fig:human eval halluci}
\end{figure*}

\begin{figure*}[!ht]
 \centering
 \includegraphics[width=1\linewidth]{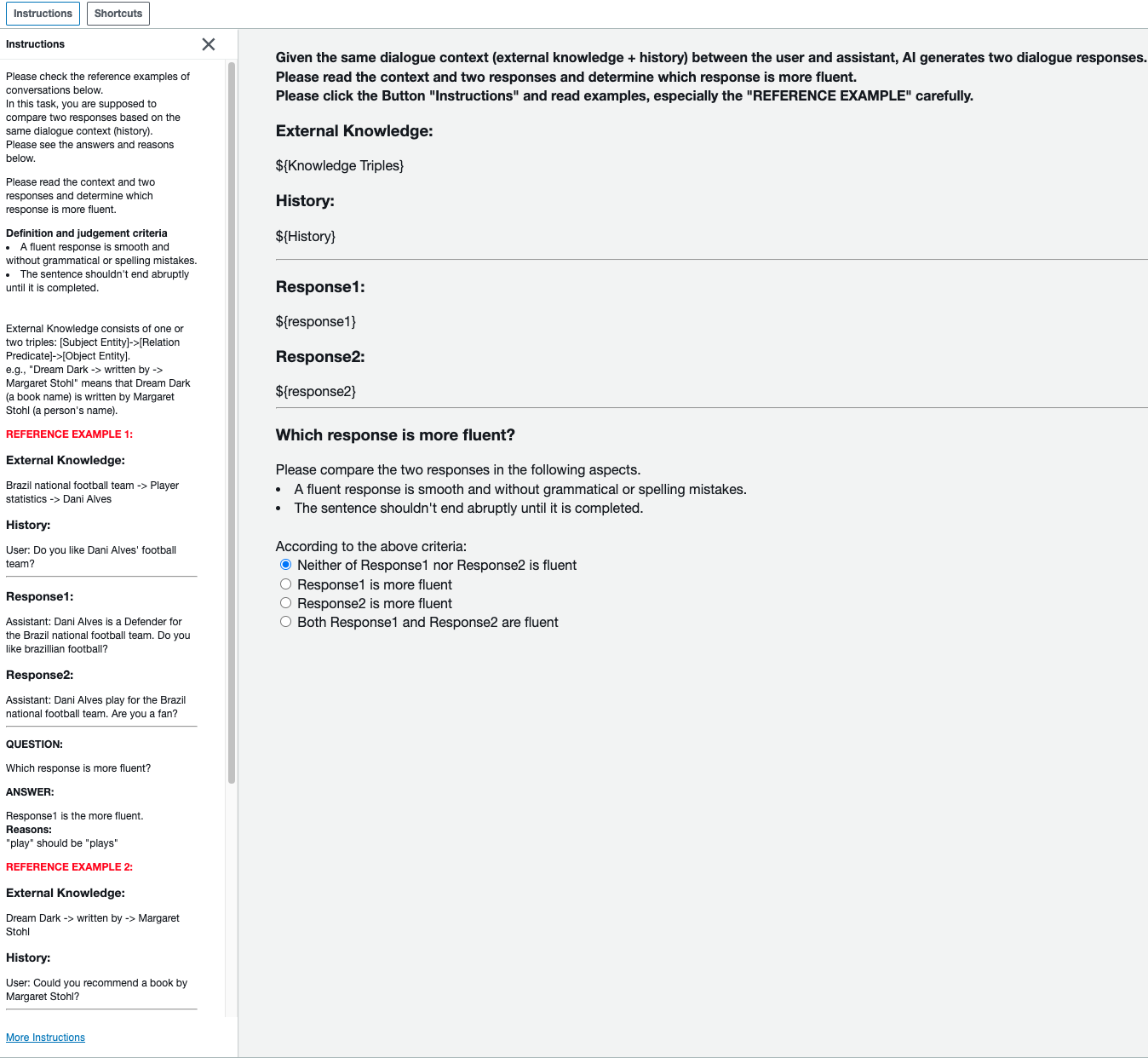}

  \caption{The UI for human evaluation on fluency via A/B testing.}
  \label{fig:human eval fluent}
\end{figure*}

We conduct the human evaluation to assess \textbf{RHO}’s performance in response generation, especially the ability to reduce hallucination.
In detail, we randomly select 100 samples generated by each model. Each sample is then evaluated by three different annotators to rule out potential bias.
We specify that annotators must meet the following qualifications: Their Human Intelligence Task (HIT) approval rates are greater than or equal to 95\%, and the numbers of HITs approved are greater than or equal to 5000. The annotators are located in Australia, the United Kingdom, and the United States. 
Figure~\ref{fig:human eval halluci} and~\ref{fig:human eval fluent} are the user interfaces (UIs) on Amazon Mechanical Turk for human evaluation of Hallucination and Fluency, respectively. The instructions, questions, and examples for annotators are shown. 

\section{KG Representation Learning}
\label{appendix sec: kg repres}
\begin{table}[]
\centering
\resizebox{1\linewidth}{!}{
\begin{tabular}{lllll}
\hline
\multicolumn{1}{c}{\multirow{2}{*}{\textbf{Model}}} & \multicolumn{2}{c}{\textbf{MR $\downarrow$}} & \multicolumn{2}{c}{\textbf{Hits@10 (\%) $\uparrow$}} \\
\multicolumn{1}{c}{}                                & \multicolumn{1}{c}{raw} & \multicolumn{1}{c}{filter} & \multicolumn{1}{c}{raw} & \multicolumn{1}{c}{filter} \\ \hline
TransE & \textbf{950.31} & \textbf{499.34} & \textbf{59.66} & \textbf{71.05}\\
TransH & 1776.69 & 946.60 & 51.71 & 67.76\\
RotaE & 1733.45 & 973.26 & 55.46 & 66.50 \\ 
SimplE & 2255.05 & 1455.30 & 45.98 & 53.21 \\ 
DistMult & 2288.26 & 1492.64 & 45.15 & 51.75 \\
ComplEx & 2381.29 & 1557.51 & 47.31 & 57.68 \\ \hline
\end{tabular}
}
\caption{Link prediction results on OpenDialKG.}
\label{tab:kg embedding}
\end{table}
For the proposed knowledge grounding techniques in \S~\ref{sec:local} and \S~\ref{sec:global}, we employ several KG representation learning algorithms including TransE~\citep{bordes2013translating}, TransH~\citep{wang2014knowledge}, RotaE~\citep{sun2019rotate}, SimplE~\citep{kazemi2018simple}, DistMult~\citep{yang2015embedding}, and ComplEx~\citep{trouillon2016complex} via OpenKE~\footnote{\url{https://github.com/thunlp/OpenKE}}.
The link prediction results are shown in Table~\ref{tab:kg embedding}, where MR is the mean rank and Hits@10 is the proportion of correct entities ranked in the top 10. 
``Filter'' means removing all corrupted triplets in the dataset~\citep{han2018openke}.
Due to the effectiveness and simplicity, we finally choose TransE to gain the KG embeddings of all entities and relation predicates in $\mathcal{G}$.

% \section{Negative Case Study}
\section{Error Analysis}
\label{appendix sec: error analysis}
\begin{table*}[!t]
\resizebox{1\linewidth}{!}{
\begin{tabular}{lllll}
\toprule
%  & \multicolumn{1}{c}{\textbf{}}
\multicolumn{1}{c}{\textbf{Category}} & \multicolumn{1}{c}{\textbf{KG}} & \multicolumn{1}{c}{\textbf{History}} & \multicolumn{1}{c}{\textbf{Golden}} & \multicolumn{1}{c}{\textbf{RHO}} \\ \hline
\begin{tabular}[c]{@{}l@{}} Extrinsic \\ Hallucination 
\end{tabular} &
\begin{tabularx}{0.18\textwidth}{X}<Windtalkers, starred actors, Nicolas Cage>\end{tabularx} &
\begin{tabularx}{0.4\textwidth}{X}User: I like American Sniper, can you recommend something similar? \\
Assistant: Sure. Windtalkers, Passchendaele, The Lying Game and The Mentalist are all in the same genre. \\
 User: Windtalkers is a Nicolas Cage flick, what genre is that exactly? \end{tabularx} &
\begin{tabularx}{0.2\textwidth}{X}Assistant: Windtalkers is a \textbf{Thriller} and a similar thriller starring him is Knowing. \end{tabularx} &         
\begin{tabularx}{0.2\textwidth}{X}Assistant: Windtalkers is a \textbf{Sci-Fi} film starring Nicolas Cage.\end{tabularx} \\ \hline

\begin{tabular}[c]{@{}l@{}} Intrinsic \\ Hallucination\end{tabular} &
\begin{tabularx}{0.18\textwidth}{X}<Todd Walker, \textbf{Drafted}, Minnesota Twins>	\end{tabularx} &
\begin{tabularx}{0.4\textwidth}{X} Assistant: Who's your favorite player in Utah Jazz? I only know of Kris Humphries. \\
User: Kris Humphries is one of my favorites! I'm just glad he doesn't play for the Minnesota Twins, although he is from there. \end{tabularx} &
\begin{tabularx}{0.2\textwidth}{X}Assistant: I'm not also fond of the Minnesota Twins. I think Todd Walker \textbf{drags} the team down.	 \end{tabularx} &         
\begin{tabularx}{0.2\textwidth}{X}Assistant: Did you know that Minnesota Twins \textbf{drafted} Todd Walker? \end{tabularx} \\ \hline

\begin{tabular}[c]{@{}l@{}} No Full\\ Coverage \end{tabular} &  
\begin{tabularx}{0.18\textwidth}{X}\textbf{<Batman, Romantically involved with, Catwoman>} \\
<Batman, starred actors, Jack Palance>\end{tabularx} &
\begin{tabularx}{0.4\textwidth}{X} Assistant: Have you seen the latest Batman movie. The Dark Knight Yet? \\
User: Not yet. Who is in it? \\
Assistant: Christopher Heyerdahl stars in The Dark Knight. Another good one is Catwoman. Have you seen it?\end{tabularx} &
\begin{tabularx}{0.2\textwidth}{X}Assistant: Who is the star in Catwoman?\end{tabularx} &         
\begin{tabularx}{0.2\textwidth}{X}Assistant: I have not seen Catwoman. Who stars in it? \end{tabularx} \\ \hline

\begin{tabular}[c]{@{}l@{}} Unnatural \\ Connection\end{tabular} &  
\begin{tabularx}{0.18\textwidth}{X}<The Rise of Theodore Roosevelt, has genre, Autobiography>	\end{tabularx} &
\begin{tabularx}{0.4\textwidth}{X} Assistant: Have you read he Rise of Theodore Roosevelt or the Bell Jar? \\
User: No, those both sound interesting though. Can you tell me more? \\
Assistant: Yes, The Rise of Theodore Roosevelt won a Pulitzer Prize for Biography or Autobiography. Would you like more modern autobiographies, as well? \end{tabularx} &
\begin{tabularx}{0.2\textwidth}{X}User: No, I think I will go with that, sounds like a winner. \end{tabularx} &     
\begin{tabularx}{0.2\textwidth}{X}User: The Rise of Theodore Roosevelt has genre Autobiography. Do you know of any other autobiographies? \end{tabularx} \\ 
\bottomrule
% \hline
\end{tabular}
}
\caption{More cases of error analysis.}
\label{tab:error}
\end{table*}

\begin{table*}[!t]
\resizebox{1\linewidth}{!}{
\begin{tabular}{lllll}
\toprule
\multicolumn{1}{c}{\textbf{KG}} & \multicolumn{1}{c}{\textbf{History}} & \multicolumn{1}{c}{\textbf{Golden}} & \multicolumn{1}{c}{\textbf{w/o. Re-ranking}} & \multicolumn{1}{c}{\textbf{w/ Re-ranking}} \\ \hline
\begin{tabularx}{0.25\textwidth}{X}<No Strings Attached, starred actors, Natalie Portman>\\
<No Strings Attached, has genre, Romantic comedy> \end{tabularx} &
\begin{tabularx}{0.15\textwidth}{X} User: Do you like Natalie Portman? \end{tabularx} &
\begin{tabularx}{0.19\textwidth}{X} Assistant: I do, Did you know she was in No Strings Attached is starring Ophelia Lovibond. It's a Romantic Comedy. \end{tabularx} &     
\begin{tabularx}{0.19\textwidth}{X}Assistant: Yes, she starred in No Strings Attached. Have you seen it? \end{tabularx} &
\begin{tabularx}{0.19\textwidth}{X}Assistant: Yes, she starred in No Strings Attached, a romantic comedy. Do you like romantic comedies? \end{tabularx} \\ 
\bottomrule
\end{tabular}
}
\caption{Case study of Re-ranking.}
\label{tab:re-rank case}
\end{table*}

As discussed, we characterize the negative cases of KDG systems into four types: Extrinsic Hallucination, Intrinsic Hallucination, No Full Coverage, and Unnatural Connection.
To gain more insights into our model, we present more failure examples from \textbf{RHO} in Table~\ref{tab:error}.

In the example of Extrinsic Hallucination, we find that in some cases, both the golden answers and our responses suffer from this issue. Based only on the given dialogue context, we cannot know or infer the genre of Windtalkers. The referenced answer is \texttt{``thriller''}, while \textbf{RHO} guesses it is  \texttt{``sci-fi''}. They are both extrinsic hallucinations since the source cannot confirm them without other information.
For Intrinsic Hallucination, the opinion in the reference answer is \texttt{``Todd Walker drags the team down''}, while \textbf{RHO} generates \texttt{``Minnesota Twins drafted Todd Walker''}. However, according to the conversation, the relationship \texttt{``Drafted''} in the KG is wrong and should be \texttt{``Dragged''}.
In the example of No Full Coverage, the first triple \texttt{``<Batman, Romantically involved with, Catwoman>''} is not mentioned in the conversation.
In the Unnatural Connection's example, our response grounds the response on the KG, but \texttt{``The Rise of Theodore Roosevelt has genre Autobiography''} is expressed mechanically and rigidly.
% and ask for other recommendations. 

\section{Response Re-ranking}

\subsection{Performance of Re-rankers}
\label{appendix sec: re-rank performance}
\begin{table}[]
\centering
\resizebox{0.9\linewidth}{!}{
\begin{tabular}{lcccc}
\hline
\multicolumn{1}{c}{\multirow{2}{*}{\textbf{Model}}}           & \multicolumn{4}{c}{\textbf{Hits@k (\%) $\uparrow$}}                                                                                 \\
\multicolumn{1}{c}{}                                          & \multicolumn{1}{c}{k=1} & \multicolumn{1}{c}{k=5} & \multicolumn{1}{c}{k=10} & \multicolumn{1}{c}{k=25} \\ \hline
\begin{tabular}[c]{@{}l@{}} w/ Sent\end{tabular} & 19.86 & 44.43 & 56.42 & 70.38 \\
\begin{tabular}[c]{@{}l@{}} w/ Sent+KG \end{tabular} & \textbf{20.49} &  \textbf{44.77} &  \textbf{56.63} & \textbf{71.47} \\   
\hline
\end{tabular}
}
\caption{Performance of re-rankers on OpenDialKG. ``Sent'' denotes using sentence embeddings, while ``Sent+KG'' denotes using both sentence and KG embeddings.}
\label{tab:re-ranker performance}
\end{table}

For entities and relations in the KG, the original settings in~\citet{sarkar2022kg} only utilize the semantic sentence embeddings via a contextual sentence encoder, i.e., Sentence-BERT.
In contrast, we sum their KG embeddings obtained by TransE with the above sentence embeddings.
The evaluation results are shown in Table~\ref{tab:re-ranker performance}, where Hits@k is the proportion of the ground-truth path ranked in the top-k highest probability paths searched by the model.
Since our settings achieve higher recall, we then use the model leveraging both sentence and KG embeddings as our re-ranker.

\subsection{Case Study of Re-ranking}
\label{appendix sec: re-rank case}
As shown in Table~\ref{tab:re-rank case}, before re-ranking, there is a No Full Coverage issue mentioned in \S~\ref{sec:error analysis}. The response does not include the information from the given triple \texttt{``<No Strings Attached, has genre, Romantic comedy>''}, which deviates from the established flow. After re-ranking, the response includes the missing information and becomes more similar to the golden answer.

% \section{Hallucination in ChatGPT}
% \label{appendix sec: chatgpt example} 
% Besides exploring ChatGPT on our task in Appendix~\ref{appendix sec: implement}, we also try other KGD generation tasks. We find that this large language model suffers from the hallucination problem, especially extrinsic hallucination. Some extrinsic hallucinations are non-factual and can raise safety issues. More exploration is needed on the detection and mitigation of hallucination in ChatGPT.

% For example 

\end{document}